\documentclass[twocolumn,10pt]{IEEEtran}
\usepackage{amsmath, amsthm, amssymb}
\usepackage{graphicx}
\usepackage{epstopdf}
\usepackage{overpic}
\usepackage{cancel}
\usepackage{rotating}
\usepackage{url}
\usepackage{caption}
\usepackage{color}
\usepackage{rotating}
\usepackage{multirow}
\usepackage{wrapfig}

\usepackage[bottom,flushmargin,hang,multiple]{footmisc}
\usepackage{lipsum}
\newcommand\blfootnote[1]{%
  \begingroup
  \renewcommand\thefootnote{}\footnote{#1}%
  \addtocounter{footnote}{-1}%
  \endgroup
}

\DeclareMathOperator*{\argmax}{arg\rm{}max}
\DeclareMathOperator*{\argmin}{arg\rm{}min}

\newcommand{\ba}{\mathbf{a}}

\newcommand{\bs}{\mathbf{s}}

\newcommand{\bw}{\mathbf{w}}

\newcommand{\bx}{\mathbf{x}}
\newcommand{\bxhat}{\mathbf{\hat{x}}}
\newcommand{\bxtilde}{\mathbf{\tilde{x}}}
\newcommand{\by}{\mathbf{y}}
\newcommand{\bz}{\mathbf{z}}
\newcommand{\bA}{\mathbf{A}}

\newcommand{\bPtilde}{\bPhitilde}

\newcommand{\bS}{\mathbf{S}}
\newcommand{\bT}{\mathbf{T}}

\newcommand{\bV}{\mathbf{V}}
\newcommand{\bX}{\mathbf{X}}
\newcommand{\bXtilde}{\mathbf{\tilde{X}}}
\newcommand{\bchi}{\boldsymbol{\chi}}
\newcommand{\bfeta}{\boldsymbol{\eta}}
\newcommand{\bmu}{\boldsymbol{\mu}}
\newcommand{\bPsi}{\boldsymbol{\Psi}}
\newcommand{\bPhi}{\boldsymbol{\Phi}}
\newcommand{\bTheta}{\boldsymbol{\Theta}}
\newcommand{\bPsitilde}{\boldsymbol{\tilde{\Psi}}}
\newcommand{\bPhitilde}{\boldsymbol{\tilde{\Phi}}}
\newcommand{\bPhihat}{\boldsymbol{\hat{\Phi}}}

\definecolor{blue}{rgb}{0,0,1}
\definecolor{darkgreen}{rgb}{0,0.5,0}
\definecolor{red}{rgb}{1,0,0}
\definecolor{teal}{rgb}{0,0.5,0.7}

\graphicspath{{./}}

\title{Optimal Sensor Placement and \\Enhanced Sparsity for Classification}
\author{Bingni W. Brunton$^{1,2*}$, Steven L. Brunton$^1$, Joshua L. Proctor$^3$, J. Nathan Kutz$^1$\\
\small{$^1$ Department of Applied Mathematics, University of Washington, Seattle, WA 98195, United States\\
$^2$ Department of Biology, University of Washington, Seattle, WA 98195, United States}\\
\small{$^3$Institute for Disease Modeling, Intellectual Ventures Laboratory, Bellevue, WA 98004, United States}\\ ~ \\
}

\begin{document}
\maketitle
\blfootnote{$^*$ Corresponding author. Tel.: +1 206 221 7991.\\ {\indent\emph{E-mail address:} bbrunton@uw.edu (B.W. Brunton).}}
%%%%%%%%%%%%
%%% ABSTRACT
%%%%%%%%%%%%
\begin{abstract}
The goal of compressive sensing is efficient reconstruction of data from few measurements, sometimes leading to a categorical decision.  
If only classification is required, reconstruction can be circumvented and the measurements needed are orders-of-magnitude sparser still.  
We define \emph{enhanced sparsity} as the reduction in number of measurements required for classification over reconstruction.  
In this work, we exploit enhanced sparsity and learn {\em spatial sensor locations} that optimally inform a categorical decision.  
The algorithm solves an $\ell_1$ minimization to find the fewest entries of the full measurement vector that exactly reconstruct the discriminant vector in feature space.  
Once the sensor locations have been identified from the training data, subsequent test samples are classified with remarkable efficiency, achieving performance comparable to that obtained by discrimination using the full image.  
Sensor locations may be learned from full images, or from a random subsample of pixels.  
For classification between more than two categories, we introduce a coupling parameter whose value tunes the number of sensors selected, trading accuracy for economy.  
We demonstrate the algorithm on example datasets from image recognition using PCA for feature extraction and LDA for discrimination; however, the method can be broadly applied to non-image data and adapted to work with other methods for feature extraction and discrimination. \\

\end{abstract}

%%%%%%%%%%%%
%%% INTRODUCTION
%%%%%%%%%%%%
\section{Introduction}
Classification and detection based on measurements is a crucial capability for intelligent systems, both biological and engineered.  
It is desirable to perform such categorical tasks with limited or incomplete information, as the full state of the system may be inaccessible or expensive to measure.  
Fortunately, even when full measurement space is quite high-dimensional, the desired information extracted from the signal is often inherently low-dimensional.  
For example, images may have a large number of pixels, but it is observed that natural images occupy a minuscule fraction of pixel space.  
Such an inherent sparsity in an appropriate transformed basis is the foundation of image compression~\cite{Elad:2010fk}; natural images may be compressed in generic Fourier or wavelet bases.

The theory of compressive sensing~\cite{Candes:2006a,Donoho:2006,Baraniuk:2007,Tropp:2007} takes further advantage of a signal's compressibility and demonstrates how a $n$-dimensional signal that is $k$-sparse in a transformed basis may be reconstructed exactly from $\mathcal{O}(k \log(n/k))$ measurements.
Here, $k$-sparse means all but $k$ coefficients are zero, and we assume $k \ll n$.    
The signal may be reconstructed from the known basis as an $\ell_0$-minimal set of sparse coefficients that best recover the measurements.  
The implementation of compressive sensing rests on the fact that the $\ell_0$-minimal  solution to an underdetermined linear system may almost certainly be found by relaxation to a convex $\ell_1$ minimization~\cite{Candes:2006a,Donoho:2006}, or by a greedy algorithm~\cite{Tropp:2007}.  

The compressive sensing strategy relies on the measurements being incoherent with respect to the known basis, so that measurement vectors are uncorrelated with basis directions.  
Incoherence holds between many pairs of bases, such as between delta functions and the Fourier basis.  Therefore, it is possible to reconstruct images, which are sparse in the Fourier domain, from single pixel measurements, which may be viewed as discrete spatial delta functions.  
Surprisingly, a random Gaussian or Bernoulli matrix is incoherent with respect to any arbitrary basis with high probability~\cite{Candes:2006a,Donoho:2006}.

Compressive sensing is able to reconstruct a signal using surprisingly few measurements, but assigning a signal to one of a few categories may be accomplished with orders-of-magnitude fewer measurements (for instance, \cite{Wright:2009}).  Typical natural images with $n$ pixels can be recovered with significant savings, usually from $n/10$ to $n/3$ measurements~\cite{Romberg:2008}.  For a 1-megapixel image, this means solving an $\ell_1$ minimization with $100,000$ constraints in a basis with one million vectors---a non-trivial computational task.  In contrast, as we will see, to classify a natural image, only tens of measurement may be required. 

Classification is often performed in a tailored low-dimensional basis extracted as hierarchical features of the data.
%A tailored low-dimensional basis extracts features from data hierarchically, often in order of energy or variance.  
One of the most ubiquitous methods in dimensionality reduction is principal components analysis (PCA)~\cite{Pearson:1901,duda:pattern}, which may be computed by a singular value decomposition (SVD) to yield an ordered orthonormal basis.  Linear discriminant analysis (LDA)~\cite{Bishop:2006fk,Fisher:1936,Rao:1948} is commonly applied in conjunction with PCA for discrimination tasks.  
Indeed, PCA-LDA is one of the classic approaches used to introduce machine learning methodologies~\cite{duda:pattern}.
High-variance PCA modes can account for common features shared across categories, which are not useful for discrimination.  
LDA produces a basis built with respect to the category geometry to maximize between-class variance while minimizing within-class variance.  PCA and LDA have both been used extensively for facial recognition (for instance, \cite{Sirovich:1987b,Turk:1991,Swets:1996,Belhumeur:1997,Yu:2001,Martinez:2001}) and is a benchmark against which novel classification techniques are often compared.

Compared to signal reconstruction, classification is particularly efficient because of 1) the use of a tailored low-dimensional feature basis, and 2) the simplicity of deciding on a category rather than reconstructing exact details~\cite{Kutz:2013}.

%%% PREVIOUS WORK
\subsection{Previous work on enhanced sparsity for classification}

Combining ideas from compressive sensing with classification, there has been significant work relevant to exploring \textit{enhanced} sparsity for classification.  
For example, the sparse representation algorithm has been applied to biometric recognition problems and demonstrated to be surprisingly robust~\cite{Wright:2009,Pillai:2011}.  
Sparse approximation applied to semantic hierarchies~\cite{Marszalek:2007} has been shown to be efficient in categorizing between large numbers of classes~\cite{Kim:2011}.  
In another line of work, ideas from compressive sensing have been applied to dimensionality reduction tools to develop a theory of sketched SVD based on randomly projected data~\cite{Fowler:2009,Qi:2012,Gilbert:2012}.  

Particularly pertinent to our work is the sparse representation for classification (SRC) algorithm as originally proposed for face recognition by Wright~\emph{et al.}~\cite{Wright:2009}.  
In SRC, $\ell_1$ minimization is used to find the sparsest representation of a test image in a dictionary composed of the training images.  Each test image is then classified based on the category of dictionary elements whose sparse coefficients best reconstructs the test image, minimizing the residual in an $\ell_2$ sense.  Notably, in this framework, the classification is robust even using dictionaries of tens or hundreds of random projections measurements, due to the enhanced sparsity of classification.

These existing algorithms have all relied on random measurements, typically using dense or sparse random measurement matrices~\cite{Gilbert:2010}.  
It remains unexplored how a refinement process may select an optimal subset of   measurements to accomplish the classification, further enhancing sparsity.  
Moreover, acquiring randomly projected measurements may be impractical where many individual point measurements are prohibitively expensive (e.g., ocean or atmospheric sensors).  
How can classification be accomplished with very few point measurements, and would the locations of those measurements illuminate coherent features of the data?

%%% PRESENT CONTRIBUTION
\subsection{Contributions and perspectives of this work}
In this work, we describe a novel framework to harness the enhanced sparsity in image recognition to classify images based on very few pixel measurements.
Its distinct contribution consists of optimally selecting, within a large set of measurement locations, a smaller subset of key locations that serve the classification.  We demonstrate that classification using very few learned pixel sensors performs comparably with using the full image.  Further, the algorithm includes a parameter to tune the trade-off between fewer sensors and accuracy. 

Datasets may be vast and data acquisition can be limited by bandwidth on data streams.  
Therefore, we also develop an intermediate technique, related to the sketched SVD~\cite{Fowler:2009,Gilbert:2010,Gilbert:2012}, to start instead with a subsample of the original data.  
We demonstrate that starting with $10\%$ of the original data, we are still able to find \emph{nearly} optimal sparse sensor locations.  

In generic applications, this entire theory is identical if single pixels are replaced with random projection measurements.  
In the case of images, the use of single pixel measurements has a number of practical advantages, since pixels are the basic unit of measurement in images.  
Note that sampling with single-pixel measurements is effective when the image features are non-localized, so that measurements and basis are incoherent.  
Importantly, sparse sensor pixels identified by our algorithm cluster near coherent features in the images.  

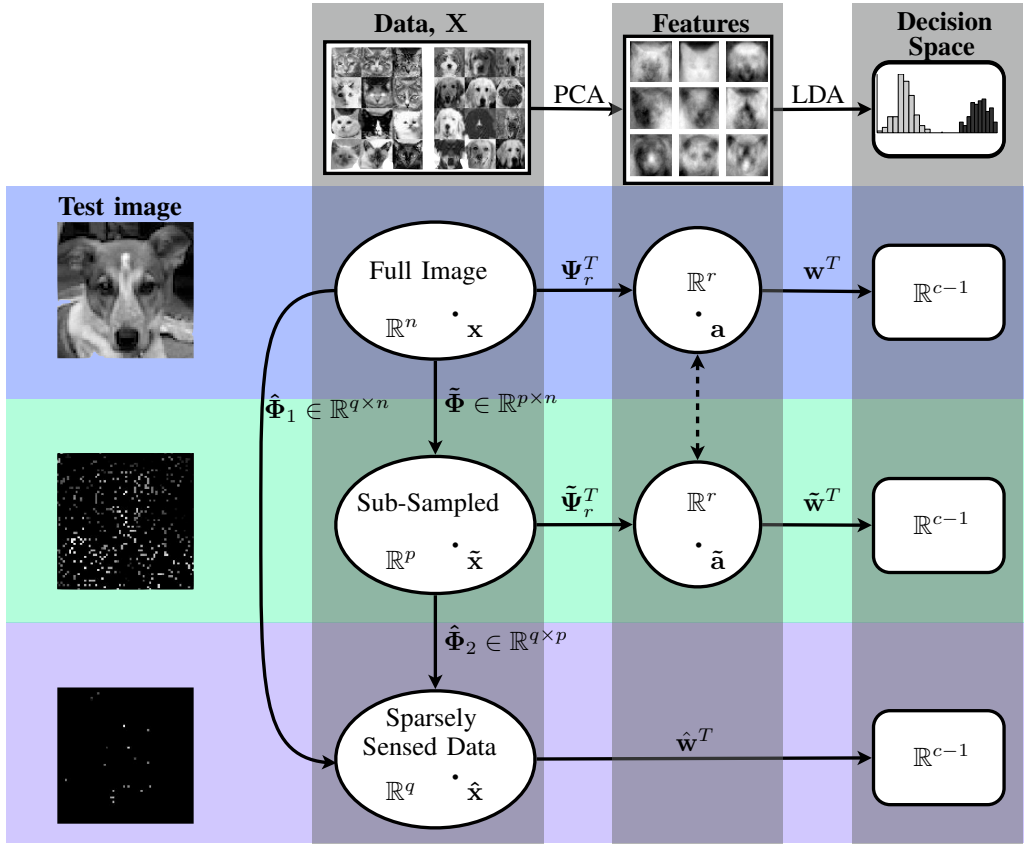
\begin{figure*}[t]
\begin{center}
\begin{overpic}[width = .85\textwidth]{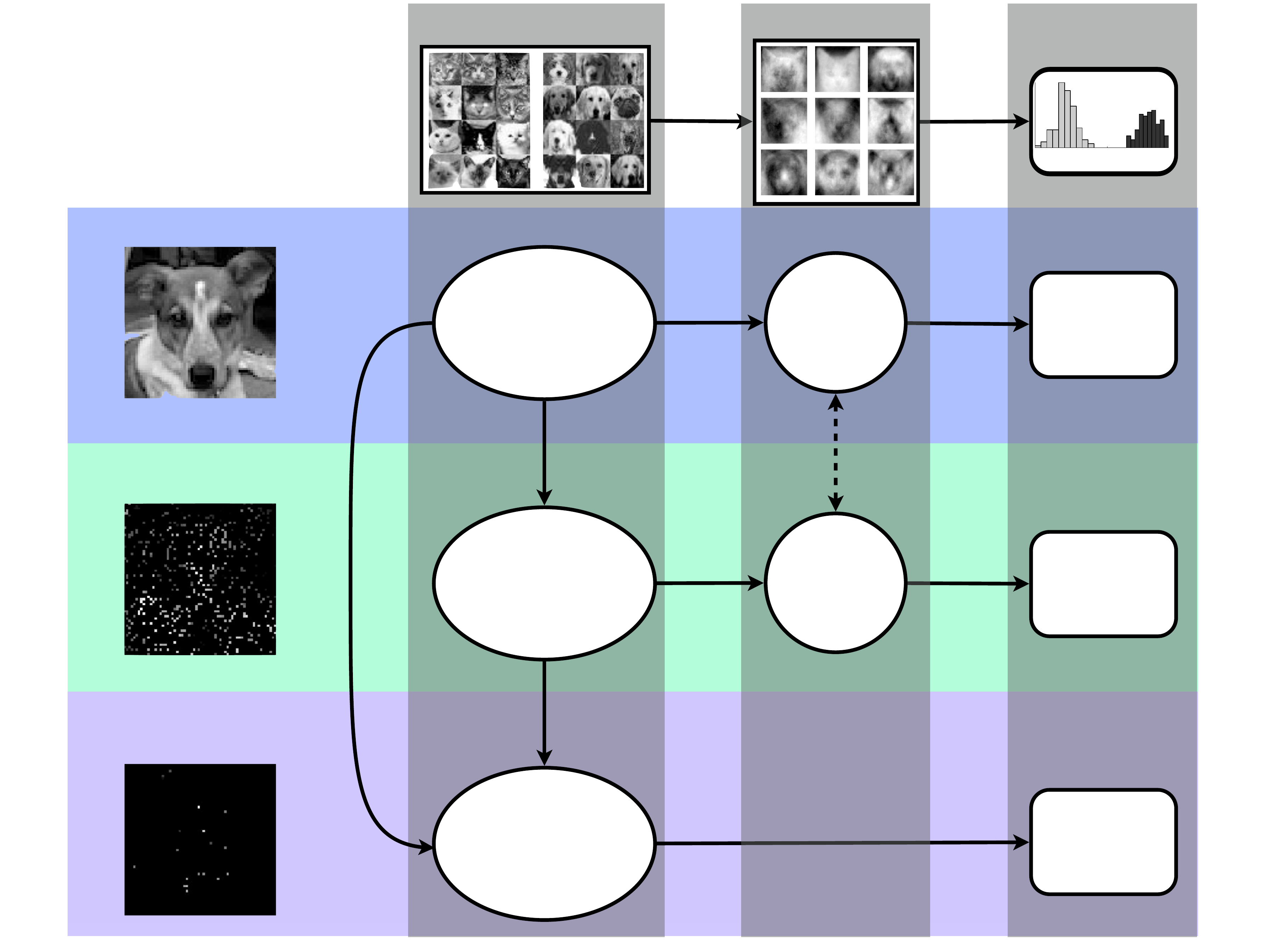}
	\put(37,70.3){\textbf{Data, $\bX$}}
	\put(61,70.3){\textbf{Features}}
	\put(82.1,70.5){\textbf{Decision}}
	\put(83.1,68.3){\textbf{Space}}
	\put(52.5,64.2){PCA}
	\put(73,64.2){LDA}
	\put(9.75,54.5){\textbf{Test image}}

	\put(36.6,49){Full Image}
	\put(38,44){$\mathbb{R}^n$}
	\put(44,46){\circle*{0.5}}
	\put(45,44){$\mathbf{x}$}
	\put(64,48){$\mathbb{R}^r$}
	\put(65,46){\circle*{0.5}}
	\put(66,44){$\ba$}
	\put(83.5,47){$\mathbb{R}^{c-1}$}
	\put(53,49){$\mathbf{\Psi}_r^T$}
	\put(74,49){$\bw^T$}

	\put(43,37.5){$\bPtilde \in \mathbb{R}^{p \times n}$}
%	\put(67.5,37.5){ref. \cite{JL:1984}}
	
	\put(35.5,29){Sub-Sampled}
	\put(38,24){$\mathbb{R}^p$}
	\put(44,26){\circle*{0.5}}
	\put(45,24){$\mathbf{\tilde{x}}$}
	\put(64,29){$\mathbb{R}^r$}
	\put(65,26){\circle*{0.5}}
	\put(66,24){$\mathbf{\tilde{a}}$}
	\put(83.5,27){$\mathbb{R}^{c-1}$}
	\put(53,29){$\mathbf{\tilde{\Psi}}_r^T$}
	\put(74,29){$\mathbf{\tilde{w}}^T$}

	\put(43,17){$\bPhihat_2 \in \mathbb{R}^{q \times p}$}

	\put(38,10.4){Sparsely}
	\put(36,7.9){Sensed Data}
	\put(38,4){$\mathbb{R}^{q}$}
	\put(44,6){\circle*{0.5}}
	\put(45,4){$\mathbf{\hat{x}}$}
	\put(83.5,7){$\mathbb{R}^{c-1}$}
	\put(63,8.3){$\hat{\bw}^T$}

	\put(27.5,37){$\bPhihat_1\in \mathbb{R}^{q \times n}$}
\end{overpic}
\vskip .2in
\caption{A schematic of the classification procedure based on full (top), randomly subsampled (middle), and sparsely sensed (bottom) data.  
Data measurements from $c$ categories (left) are projected onto a $r$-dimensional PCA feature space (middle).  LDA defines the projection $\mathbf{w}^T$ from PCA space to $\mathbb{R}^{c-1}$ (right), where classification occurs.  
The $\bPtilde$ matrix is a random projection that sub-samples the full image; the $\bPhihat_1$ matrix is a learned projection that samples the full image at specific sensors as described in Sec.~\ref{sec:methods}.  Sparse sensors can also be learned from the randomly sub-sampled image, resulting in $\bPhihat_2 \bPtilde$, which is not the same as, but may approximate, $\bPhihat_1$.  Typically, $q \ll p \ll n$. The number of sensors $q$ is at most $\bar{q}$, which is bounded by $r\leq \bar{q}\leq r(c-1)$; for two-way classification ($c=2$), $\bar{q}=r$.}
\label{fig:schematic}
\end{center}
\end{figure*}

Our framework for sparse classification has two characteristics that distinguish it significantly from previous work: 1)~instead of random measurements, spatial sensor locations are specifically selected; and 2) $\ell_1$ minimization is applied {\em once} in sensor learning, and classification of new images involves dot products on a very small measurement vector.
A schematic representation of our procedure is found in Fig.~\ref{fig:schematic} with details following in the text.  
Equations~\eqref{eq:ell1} and \eqref{eq:ell1extended} comprise our primary
theoretical contributions; Figs.~\ref{fig:ell1} and \ref{fig:sensor_locations_lambda} illustrate the sensor locations produced by the algorithm.  

This work takes an engineering perspective of probing complex systems with underlying low-dimensional structure, with the explicit goal of forming some categorical decision about the state of the system.
The optimally sparse spatial sensors framework is particularly well suited for engineering applications, as an upfront investment in the learning phase allows remarkably efficient performance for all subsequent queries.  Abstractions of this method to more general data sets is discussed in more detail in Sec.~\ref{sec:discussion}.

We speculate that the principle of enhanced sparsity may also have relevance for biological organisms, which often need to make decisions based on very limited sensory information.  
Specifically, organisms interact with high-dimensional physical systems in nature but must rely on information gathered through a handful of sensory organs.  
Our sparse sensor algorithm provides one approach to answer the question, Given a fixed budget of sensors, where should they be placed to optimally inform decision-making?

%%% ORGANIZATION
\subsection{Organization of the paper}
Section~\ref{sec:background} provides an overview of well-known techniques on which we build our contributions and establishes the notation used for the remainder of the paper.  We summarize compressive sensing, random projections for matrix decomposition, sparse representation for classification, and the PCA-LDA method for categorization.
Section~\ref{sec:methods} describes our algorithm for determining optimally sparse sensor locations by using $\ell_1$ minimization and a dictionary of learned features.  
We applied sensor learning to two image discrimination tasks based on real-world data; the results of these experiments are presented in Sec.~\ref{sec:results} and the methods are discussed more generally in Sec.~\ref{sec:discussion}.

\section{Background}
\label{sec:background}
The main contributions of this work are 1) to design optimal sensor locations that take advantage of enhanced sparsity in classification, and 2) to design sensors from substantially sub-sampled data.  
To put our contributions in context, we review well known results from compressive sensing in Sec.~\ref{ss:CS}, which describes the theory of $\ell_1$ reconstruction, sparse random projection matrices, and sparse representation for classification.

We also make use of two well-established methods for systematically producing low-rank representations of a high-dimensional data set, principal components analysis (PCA) and linear discriminant analysis (LDA).  
In this manuscript, we demonstrate our algorithm by applying PCA in combination with LDA for face recognition.  
In general, the method can be implemented with many other dimensionality-reduction and discrimination algorithms, as drawn schematically in the top of Fig.~\ref{fig:schematic}.  
Section~\ref{ss:PCALDA} reviews PCA-LDA and establishes the notation used to describe our methods in Sec.~\ref{sec:methods}.

%%% COMPRESSIVE SENSING
\subsection{Compressive sensing and sparse representation}
\label{ss:CS}
Compressive sensing theory states that if the information of a signal $\bx \in \mathbb{R}^n$ is $k$-sparse in a transformed basis $\bPsi$ (all but $k$ coefficients are zero) and $k \ll n$, it is possible to reconstruct the signal from very limited measurements $\mathcal{O}(k \log(n/k))$~\cite{Candes:2006, Candes:2006a, Candes:2008,Candes:2010}.  
This technique has widespread applications, including the fields of medical imaging~\cite{Lustig:2008}, neuroscience~\cite{Ganguli:2012}, and engineering~\cite{Herman:2009}.  

The standard framing of the compressive sensing problem is as follows.  Let us suppose the measurement vector $\bxtilde \in \mathbb{R}^p$ is related to the desired signal $\mathbf{x} \in \mathbb{R}^n$ by $\bPhitilde$, a known $p\times n$ measurement matrix: $\bxtilde = \bPhitilde \bx$.  Compressive sensing seeks to recover $\bx$ from $\bxtilde$ and applies to the underdetermined case where $p \ll n$.

The signal $\bx$ has coefficients $\ba$ in basis $\bPsi$, so that
\begin{align}
\bxtilde = \bPhitilde\bx=\bPhitilde\bPsi\ba = \bTheta\ba.
\label{eq.CS}
\end{align}
If $\bx$ is sufficiently sparse in $\bPsi$ and the matrix $\bTheta$ obeys the restricted isometry principle (RIP)~\cite{Candes:2005}, the search for $\ba$ (and thus the reconstruction of $\bx$) is possible.  

We would like to solve for the sparsest $\ba$,
\begin{align}
\ba = \argmin_{\ba'} \|\ba'\|_0, \text{~~subject to}~\bxtilde = \bTheta\ba',
\label{eq:csell0}
\end{align}
but this involves an intractable combinatorial search.  It has been shown that under certain conditions, Eq.~\eqref{eq:csell0} may be solved by minimizing the $\ell_1$ norm \cite{Candes:2006a,Donoho:2006b},
\begin{align}
\ba = \argmin_{\ba'} \|\ba'\|_1, \text{~~subject to}~\bxtilde = \bTheta\ba'.
\label{eq:csell1}
\end{align}
Relaxation to a convex $\ell_1$ minimization bypasses the combinatorially difficult problem.  Solutions to Eq.~\eqref{eq:csell1} may be found through standard convex optimization routines, or by greedy algorithms such as orthogonal matching pursuit~\cite{Tropp:2006b,Tropp:2007}.  

An important aspect of compressive sensing is the choice of the measurement matrix $\bPhitilde$.  
To achieve exact reconstruction, $\bPhitilde$ must be incoherent with respect to $\bPsi$ (therefore satisfying the RIP).  Many authors describe the properties of random matrix ensembles that fulfill these conditions with high probability: they include Gaussian, Bernoulli, and random partial Fourier matrices~\cite{Candes:2006, Candes:2006b, Donoho:2006}.  In addition, sparse random matrices, where many of the entries of the measurement vectors are zero, also allow reconstruction (as reviewed in \cite{Gilbert:2010}).  Theoretical limits of how sparse random projections may be and still support signal recovery have been explored~\cite{Li:2006, Wang:2010}.

In addition to the random measurement framework for reconstruction, we are concerned with the related question of the spectral properties of a data matrix under random projection.  Consider a data matrix $\bX=\begin{bmatrix}\bx_1&\hskip-.03in \bx_2& \hskip-.04in \dots& \hskip-.04in\bx_m \end{bmatrix}\in\mathbb{R}^{n\times m}$.  Let
\begin{align*}
\bXtilde = \bPhitilde \bX,
\end{align*}
where $\bPhitilde \in \mathbb{R}^{p \times n}$ and $p \ll n$, as above.  
%Under what conditions are the spectral properties (i.e., the singular values and vectors) of $\bXtilde$ and $\bX$ approximately equal?  Gilbert \emph{et al.}~\cite{Gilbert:2012} found that 
Spectral properties of $\bX$ and $\bXtilde$ are approximately equal when the random measurement matrix $\bPhitilde$ satisfies the distributed Johnson-Lindenstrauss (JL) property~\cite{JL:1984, Gilbert:2012}.  This property regards the preservation of relative distances between points after random projection to a new space.  
Since the classification task depends on a geometric separation of data points between classes, the JL property is integral to the success of this model reduction step (dashed arrow between the full and subsampled PCA feature spaces in Fig.~\ref{fig:schematic}).

A related line of work on sparse representation for classification (SRC)~\cite{Wright:2009} proposes to leverage sparsity promotion by $\ell_1$ minimization for image recognition.  In this formulation of the classification problem, each training image makes up a column of the dictionary $\bTheta$, and the test image $\by$ is represented as a linear combination of the dictionary elements weighted by coefficients $\ba$.  Thus posed, the solution to the following equation is the sparsest representation of the test image given the dictionary,
\begin{align*}
\ba = \argmin_{\ba'} \|\ba'\|_1, \text{~~subject to}~\by = \bTheta\ba'.
\end{align*}
Note that the above equation is identical to Eq.~\eqref{eq:csell1}.

It is then possible to construct a sparse approximation of $\by$ using only coefficients in $\ba$ associated with the $i$th class,
\begin{align*}
\mathbf{\tilde{y}}_i = \bTheta \delta_i(\ba_i),
\end{align*}
where $\delta_i(\ba)$ is a vector the same size as $\ba$ whose only non-zero entries are the entries in $\ba$ associated with training images in class $i$.  Finally, $\by$ is assigned a category based on the class whose approximation minimizes the residual between $\by$ and $\mathbf{\tilde{y}}_i$:
\begin{align*}
\text{category~}(\by) = \argmin_i || \by - \bTheta \delta_i(\ba_i) ||_2.
\end{align*}

The SRC framework exploits enhanced sparsity for classification, making use of the sparse subspace structure of face images. It has been demonstrated to work exceptionally well for many possible dictionaries $\bTheta$, including common features such as eigenfaces and Fisher faces, and unncommon features including downsampled faces and random projections.

It is worth noting that the performance of SRC for face recognition is achieved at the cost of one $\ell_1$ minimization for each test image.  
As we will see in Sec.~\ref{sec:methods}, this is distinct from our algorithm, which uses $\ell_1$ minimization to identify sparse sensor locations in the training phase only. 

%%% PCA-LDA
\subsection{PCA, LDA, and classification}
\label{ss:PCALDA}
Here we describe the two main dimensionality reduction techniques used in this manuscript to demonstrate the pixel refinement algorithm in Sec.~\ref{sec:methods}.  
Consider a set of observations $\bx_i$, $i=1,\dotsc,m$, where $\bx_i \in \mathbb{R}^n$; also suppose that $m\ll n$.
Let us construct a data matrix $\bX$ with columns $\bx_i$, as represented schematically in Fig.~\ref{fig:schematic}; we assume that rows of $\bX$ have been mean subtracted. The PCA consists of performing a singular value decomposition (SVD) defined by the following modal decomposition:
\begin{align}
\bX=\bPsi\mathbf{\Sigma} \bV^*.
\label{eq.SVD}
\end{align}

The columns of $\bPsi$ are the left singular vectors of $\bX$; they span the columns of $\bX$ and are often referred to as the \emph{principal components} or \emph{features} of the data set.  The matrix $\mathbf{\Sigma}$ is diagonal; entries $\sigma_{ii}$ are the \emph{singular values} of $\bX$.  There are only $m$ non-zero singular values, and we may write $\mathbf{\Sigma}=\begin{bmatrix}\mathbf{\Sigma}_{m\times m} & \mathbf{0}\end{bmatrix}^T$.  

The columns of the matrix $\bPsi$ are eigenvectors of $\bX\bX^T$:
\begin{align*}
\bX\bX^T\bPsi = \bPsi \mathbf{\Lambda}, \quad \mathbf{\Lambda} = \begin{bmatrix}\mathbf{\Sigma}_{m\times m}^2& \mathbf{0} \\ \mathbf{0} & \mathbf{0}\end{bmatrix}.
\end{align*}
However, this is an extremely expensive eigendecomposition, since the dimension of $\bX\bX^T$ is $n\times n$.  In implementation, it is more practical to use the method of snapshots~\cite{Sirovich:1987}, whereby we solve the $m\times m$ eigendecomposition for $\bV$:
\begin{align*}
\bX^T\bX \bV = \bV \mathbf{\Sigma}_{m\times m}^2.
\end{align*}
We may then obtain the $m$ non-zero PCA modes by taking a linear combination of the columns of $\bX$:
\begin{align*}
\bPsi_m = \bX \bV \mathbf{\Sigma}_{m\times m}^{-1}.
\label{eq:snapshotPCA}
\end{align*}

The SVD provides a systematic approach to transform data into a lower-dimensional representation.  
For data sets that are inherently low-rank, the singular values $\mathbf{\Sigma}$ of Eq.~(\ref{eq.SVD}) contains many entries that are exactly zero.
With most realistic applications and data sets, though, the singular values often exhibit a power-law decrease, where the low-rank representation of the data is approximate.  
A power-law decrease of singular values provides an opportunity for a heuristic, and at least quantitatively informed, decision about the number of information-rich dimensions to retain for the reduced-order subspace.   
Taking the $r$ columns of $\bPsi$ as basis vectors corresponding to the $r$ largest singular values, a truncated basis $\bPsi_r$ can be formed.  
We may use this basis to project data from the full measurement space into the reduced $r$-dimensional PCA space (known as \emph{feature space}):
\begin{align}
\begin{split}\bPsi_r^T:\mathbb{R}^n&\rightarrow \mathbb{R}^r,  \\
\bx&\mapsto \ba.
\end{split}
\end{align}
Thus, the principal components $\bPsi_r$ minimize the $\ell_2$ projection error, $\|\bx-\bPsi_r\bPsi_r^T\bx\|_2$.

For categorical decisions, we apply the well known classifier linear discriminant analysis (LDA) in $r$-dimensional feature space.  
Despite the plethora of classifier techniques, we chose LDA for its simplicity and well-established behavior.  
We note once again that PCA and LDA are serving only to illustrate the concept proposed in Fig.~\ref{fig:schematic}.

We start by noting that each observation $\bx_i$ (and thus each $\ba_i=\bPsi_r^T\bx_i$) belongs to one of $c$ distinct categories $C_j$ where $j = 1, \dotsc, c$.
LDA attempts to find a set of directions in feature space $\bw \in \mathbb{R}^{r \times c-1}$ where the between-class variance is maximized and within-class variance is minimized.  
Specifically,
\begin{align}
\bw = \argmax_{\bw'} \frac{\bw'^T \bS_B \bw'}{\bw'^T \bS_W \bw'}.
\end{align}
Here $\bS_W$ is the within-class scatter matrix,
\begin{align*}
\bS_W = \sum_{j=1}^c \sum_{i \in C_j} (\ba_i - \bmu_j) (\ba_i - \bmu_j)^T,
\end{align*}
where $\bmu_j$ is the mean of class $j$.  $\bS_B$ is the between-class scatter matrix,
\begin{align*}
\bS_B = \sum_{j=1}^c N_j (\bmu_j - \bmu) (\bmu_j - \bmu)^T,
\end{align*}
where $N_j$ is the number of observations in class $j$ and $\bmu$ is the mean of all observations in $\bA=\bPsi_r^T\bX$ (in this case $\bmu=0$).  

It follows that $\bw$ are eigenvectors corresponding to the non-zero eigenvalues of $\bS_W^{-1}\bS_B$.
It should be noted that the number of non-trivial LDA directions must be less than or equal to the number of categories minus one ($c-1$), and that at least $r+c$ samples are needed to guarantee $\bS_W$ is not singular~\cite{Bishop:2006fk}.  

In schematic form, the top ``full image'' row of Fig.~\ref{fig:schematic} illustrates the PCA-LDA process for the classification task.  
In the case of $c = 2$ categories, the LDA projection is $\bw^T:~\mathbb{R}^r~\rightarrow~\mathbb{R}$, and we may apply a threshold value that separates the two categories.  
Assuming the two categories have equal covariances, we pick the threshold to be the midpoint between the means of the two classes, $\bw^T \bmu_A$ and $\bw^T \bmu_B$. 

For decisions between $c \ge 2$ categories, LDA produces a projection $\bw^T: \mathbb{R}^r \rightarrow \mathbb{R}^{c-1}$.  
We use a nearest centroid (NC) method for classification, in which a new measurement $\bx_i$ is assigned to category $j$ for which the distance between $\bw^T \bx_i$ and $\bw^T \bmu_j$ is minimal.  
Many other classifier choices are possible in the reduced $\mathbb{R}^{c-1}$ space, including nearest neighbor (NN)~\cite{duda:pattern}, nearest subspace (NS)~\cite{Lee:2005fk}, support vector machines (SVM)~\cite{Munoz:2006uq}, and sparse representation (SRC)~\cite{Wright:2009}.  
While it is possible these classifiers may improve the accuracy of the decision, we restricted our attention to NC since the focus of this work is not on the specific decision algorithm, but rather on the method of learning sensor locations.

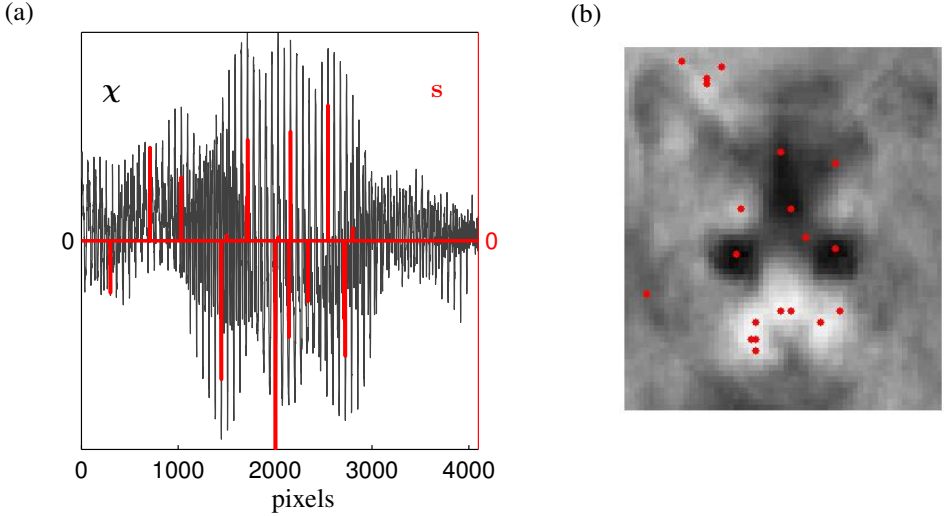
\begin{figure*}[tbh]
\begin{center}
\begin{tabular}{cc}
\begin{overpic}[width=0.375\textwidth]{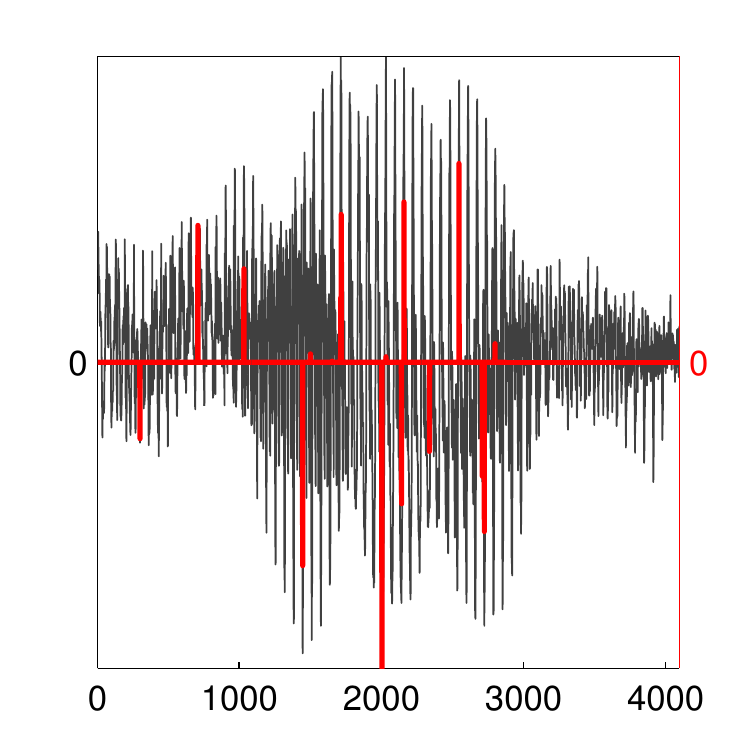}
\put(-2,95){(a)}
\put(50,0){pixels}
\put(17,80){$\bchi$}
\put(81,80){\color{red} $\bs$}
\end{overpic}&
\begin{overpic}[width=0.375\textwidth]{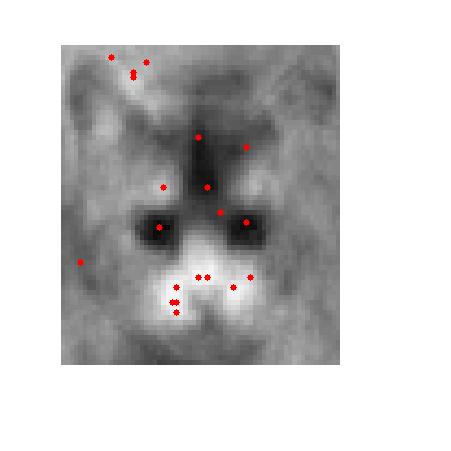}
\put(3,95){(b)}
\end{overpic}
\end{tabular}
\caption{A visualization of $\bchi\triangleq \bPsi_r \bw$ and how sparse sensors, as described in Sec.~\ref{sec:methods}, approximate its dominant features. (a) $\bchi$ (black) and sparse approximation $\bs$ (red), where $r = 20$.  (b) An alternative visualization of $\bchi$ (grey) and locations of sparse sensors (non-zero elements of $\bs$, in red).  See Sec.~\ref{sec:results:catdog} for specific details of cat/dog image recognition.}
\label{fig:ell1}
\end{center}
\end{figure*}

%%%%%%%%%%%%
%%% METHODS (SPARSE SENSORS)
%%%%%%%%%%%%
\section{Obtaining sparse sensor locations}\label{sec:methods}
This section describes our method to learn sparse sensor locations, building upon the theory described in Sec.~\ref{sec:background}.  
We begin in Sec.~\ref{ss:twowaysensors} by addressing classification between $c=2$ categories, reducing a full measurement vector to a small handful of key measurements.  
Section~\ref{ss:randomsubsample} describes an intermediate technique that starts instead with a random subsample of the full data.
In Sec.~\ref{ss:multiwaysensors} we develop an extension of the algorithm to classification between $c \ge 2$ categories.  
This subsection also introduces a coupling weight $\lambda$, which allows us to decrease the number of sensors required at the cost of slightly lower classification accuracy.  
Finally, Sec.~\ref{ss:ldaprojection} demonstrates the possibility of transforming the full-dimensional feature/discrimination projection into an approximate projection from the learned sparse measurements to the decision space.  
Alternatively, it is possible to re-compute a new projection to decision space directly from the data matrix comprised of sparse measurements, as in the last row of Fig.~\ref{fig:schematic}.  

In the following, we refer to \emph{features} $\bPsi_r$ (e.g., PCA modes) and \emph{discrimination vectors} $\bw$ (e.g., LDA directions) to emphasize the generality of the method.

%%% TWO CATEGORIES
\subsection{Deciding between two categories} \label{ss:twowaysensors}

Let us first consider learning sparse sensors for a classification problem between $c$~=~2 categories.  
The discrimination vector $\bw$ encodes the direction in $\bPsi_r$ that is most informative in discriminating between categories of observations given by $\bX$.  
We seek a measurement vector $\bs$ that satisfies $\bPsi_r^T\bs=\bw$.  

In particular, we seek the sparse solution $\bs$:
to find the measurements that best reconstruct $\bw$, our approach is to solve for
\begin{align}
\bs = \argmin_{\bs'} ||\bs'||_1, \text{~~subject to}~\bPsi_r^T \bs' = \bw.\label{eq:ell1}
\end{align}

The equations for $\bs$ are under constrained, so that we may use $\ell_1$ minimization to find the sparse solution to this convex problem.
As we will show in Sec.~\ref{sec:results}, the solution for $\bs$ has at most $r$ nonzero elements.  Alternatively, it is possible to solve Eq.~\eqref{eq:ell1} for exactly $r$ nonzero elements using a greedy algorithm~\cite{Tropp:2007}.

It is important to remember that $\bs \in \mathbb{R}^n$ and can be visualized as an image where most of the pixels are zero.  
The $r$ non-zero elements of $\bs$ represent locations of sensors that best recapture the discriminant projection vector $\bPsi_r\bw$. 

To motivate this approach, consider that we have constructed a projection $\bPsi_r^T$ onto a set of features and a projection $\bw^T$ from feature space to decision space; we may simplify this into a single projection:
\begin{align*}
\eta&=\bw^T\bPsi_r^T\bx_{\text{test}}\\
&=\langle \bchi, \bx_{\text{test}}\rangle
\end{align*}
where $\bchi\triangleq \bPsi_r \bw$.  Therefore, Eq.~\eqref{eq:ell1} finds the sparse measurement vector $\bs$ that projects to the same feature space coordinates as $\bchi$:
\begin{align*}
\bPsi_r^T\bs &= \bPsi_r^T \bchi\\
&=\underbrace{\bPsi_r^T\bPsi_r}_{\mathbb{I}^{r\times r}} \bw
=\bw
\end{align*}
Finally, we see that
\begin{align}
\bs=\bchi + \boldsymbol{\xi}.
\end{align}
where $\boldsymbol{\xi}\in\ker(\bPsi_r^T)$.
That is to say, the sparse measurement $\bs$ is the same as the vector $\boldsymbol{\chi}$ plus some residual vector that does not contain any dominant features (i.e., is not in the span of dominant features).

To compare $\bchi$ and $\bs$, Fig.~\ref{fig:ell1} visualizes these vectors for the dog/cat face recognition problem discussed in Sec.~\ref{sec:results:catdog}. 
Figure~\ref{fig:ell1}\,(a) shows the magnitude $\bchi$ for $r = 20$ along with the 20 pixel measurement vector $\bs$ obtained by $\ell_1$ minimization of Eq.~(\ref{eq:ell1}).
The image of $\bchi$ (grey) and the sparse pixels (red) are shown in Fig.~\ref{fig:ell1}\, (b).  
Importantly, it is not possible to obtain sparse pixel measurements by thresholding $\bchi$ (Fig.~\ref{fig:ell1}\,(a)); rather, $\bs$ is a sparse image that exactly projects to $\bw$ in $\bPsi_r$ space.

To implement this optimal sub-sampling at learned sensors locations, we construct a $q \times n$ projection matrix $\bPhihat_1$ that maps $\bx \mapsto \mathbf{\hat{x}}$.  
The number of sensors $q$ is usually equal to $r$.  
The matrix $\bPhihat_1$ are rows of the $n \times n$ identity matrix corresponding to the non-zero elements of $\bs$.

%%% SUBSAMPLED DATA
\subsection{Learning from randomly subsampled data} \label{ss:randomsubsample}

The identical approach applies when starting from sub-sampled data $\mathbf{\tilde{X}}$ (the middle ``sub-sampled'' row of Fig.~\ref{fig:schematic}).  
We now solve for $\mathbf{\tilde{s}}$ that best reconstructs the vector $\mathbf{\tilde{w}}$ in a space defined by the columns of $\bPsitilde_r$.  
$\bPtilde$ is then refined to construct $\bPhihat_2$, the $r$ rows of the $p \times p$ identity matrix corresponding to non-zero elements of $\mathbf{\tilde{s}}$.

Note that $\bPhihat_1$ and $\bPhihat_2 \bPtilde$ are both $r \times n$ matrices, where typically $r \ll p \ll n$.  
Although the two sub-sampling matrices are not the same, we show that in Sec.~\ref{sec:results} that they are quite similar when $p/n \approx 0.1$. 

Finally, having learned sparse sensor locations, we project the data into $\mathbb{R}^r$,
\begin{align*}
\hat{\bX} = \bPhihat \bX,
\end{align*}
\noindent where $\bPhihat$ can be either $\bPhihat_1$ or $\bPhihat_2 \bPtilde$, as illustrated in Fig.~\ref{fig:schematic}.  
Possible methods to project these learned sparse measurements to decision space are described in Sec.~\ref{ss:ldaprojection}.

%%% MORE THAN TWO CATEGORIES
\subsection{Deciding between more than two categories} \label{ss:multiwaysensors}
The simplest extension to classification between $c \ge 2$ categories can be implemented by considering the projection $\bw^T:\mathbb{R}^r \rightarrow \mathbb{R}^{c-1}$ to decision space, followed by independently solving Eq.~\eqref{eq:ell1} for each column of $\bw$.  
However, this approach scales badly with $c$; in general, discriminating $c$ categories of data by projection into $r$-dimensional feature space results in at most $q = r(c-1)$ learned sensors locations.

An alternative formulation of the convex optimization problem solves for columns of $\bs \in \mathbb{R}^{n\times (c-1)}$ simultaneously; each column of $\bs$ (image) projects to a column of $\bw$ (discrimination vector) in feature space.  
We introduce a norm that penalizes the total number of non-zero rows in $\bs$ (pixel measurements) to reconstruct the $c-1$ columns of $\bw$.  
Specifically,
\begin{align}
\bs = &\argmin_{\bs'} \left\{ || \bs' ||_1 + \lambda || \bs' \mathbf{v} ||_1 \right\}, \nonumber \\
&\text{subject to}~ || \bPsi_r^T \bs' - \bw ||_F \leq \varepsilon, \label{eq:ell1extended}
\end{align}
where $||\mathbf{M}||_1 = \sum_{ij} |m_{ij}|$, $\mathbf{v}$ is a column vector of $(c-1)$ ones, $||\mathbf{M}||_F = \sqrt{\sum_{ij} |m_{ij}|^2}$ is the Frobenius norm, and $\varepsilon$ is a small error tolerance ($\varepsilon\approx 10^{-10}$ for examples in Sec.~\ref{sec:results}).

Once again, we have an underdetermined system of equations for $\bs$.  
The value of the coupling weight $\lambda$ determines the number of non-zero rows of $\bs$, so that the number of sensors $q$ identified is at most $\bar{q}$ where $r \le \bar{q} \le r(c-1)$.

In the limit where $\lambda = 0$, the solution to Eq.~\eqref{eq:ell1extended} is the same as obtained by the uncoupled, independent approach.  
In general, solving Eq.~\ref{eq:ell1extended} with $\lambda~=~0$ would result in at most $r(c-1)$ sensor locations.  
As $\lambda$ becomes larger, the coupling between columns of $\bs$ becomes stronger, and the same pixel location can be shared between columns of $\bs$ to approximately reconstruct columns of $\bw$.  
In other words, the same pixel measurement is re-used to capture multiple linear discriminant projection vectors.  
In the limit $\lambda~\rightarrow~\infty$, the number of sensors is bounded $r$.

The optimization problem as formulated in Eq.~\eqref{eq:ell1extended} is closely related to the one solved by Simultaneous Orthogonal Matching Pursuit (S-OMP,~\cite{Tropp:2006b, Tropp:2006c}).  
S-OMP is a greedy algorithm, so instead of a coupling weight parameter $\lambda$, one would decide on a stopping criterion (for example, the desired number of iterations/sensors).

Following the same approach as in deciding between two categories, we implement sub-sampling at learned sensors to construct a projection matrix $\bPhihat$.  

%%% PROJECTION TO DECISION SPACE
\subsection{Projecting sparse measurements to classification space} \label{ss:ldaprojection}
In the above procedure, the projections from the high-dimensional space into feature space ($\bPsi_r^T$) and then into decision space (${\bf w}^T$) are computed as a one-time upfront cost.  
It is then possible to re-use these projections to obtain an induced projection into the decision space starting only from learned sparse measurements.  
The alternative is to re-compute the discrimination vectors on the sparse measurement data $\bf\hat X$.  

Given a new image $\bx$, we obtain a vector $\bfeta$ of decision space coordinates as follows:
\begin{equation}
\bfeta= \bw^T\bPsi_r^T\bx. 
\label{eq:reLDA1}
\end{equation}
We then substitute $\bw=\bPsi_r^T\bs$ and $\bx\approx \bPhihat^T\bxhat$, where $\bxhat=\bPhihat \bx$, into Eq.~\eqref{eq:reLDA1}:
\begin{equation}
\tilde{\bfeta} = \bs^T\bPsi_r\bPsi_r^T\bPhihat^T\bxhat.
\label{eq:reLDA2}
\end{equation}
where $\tilde{\bfeta}\approx \bfeta$.  
Define $\bz$ as the measurement projection of $\bs$, so that $\bz=\bPhihat\bs$ and $\bs\approx \bPhihat^T\bz$.  
Substituting into Eq.~\eqref{eq:reLDA2} yields:
\begin{align*}
\tilde{\bfeta}&=\bz^T\bPhihat\bPsi_r\bPsi_r^T\bPhihat^T\bxhat \\
&=\bz^T\bT \bxhat. 
\end{align*}
The matrix $\bT=\bPhihat\bPsi_r\bPsi_r^T\bPhihat^T$ defines a new inner-product on the space of sparse measurements that preserves the geometry of decision space.  
To compute $\bT$ efficiently, compute $\bPhihat\bPsi_r$ and $\bPsi_r^T\bPhihat^T$ separately and then multiply the results.  

The alternative approach of re-computing the discrimination projection on the sparse data ${\bf\hat X}=\bPhihat \bX$ is typically inexpensive because of the small number of rows.  
Additionally, for multi-way classification, this usually leads to better performance, as discussed in Sec.~\ref{sec:results}.

%%%%%%%%%%%%
%%% RESULTS
%%%%%%%%%%%%
\section{Experiments}\label{sec:results}
We apply the algorithm for learning sparse sensor locations, as describe in Sec.~\ref{sec:methods}, on two examples of face recognition based on publicly available image datasets.  
Both data sets are presented and described in Appendix~\ref{app:datasets}.
In each experiment, we demonstrate that a classifier built on a few optimally placed sparse sensors performs comparably to a classifier built on PCA features of the full image.  
Moreover, approximately optimal sparse sensor locations may be learned from pixels randomly subsampled from 10\% of the full image.  
Ensembles of the learned sensors cluster at locations consistent with the coherent features in the faces.

\begin{figure}[h]
\begin{center}
\begin{overpic}[width=0.43\textwidth]{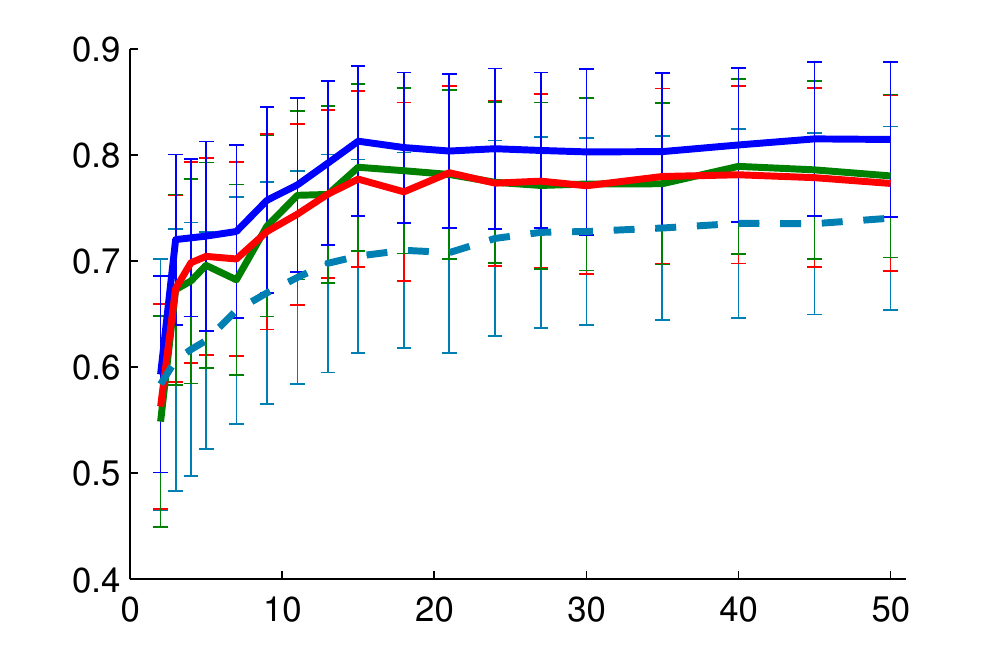}
	\put(0,-5){$r$, Number of features/sensors used in classifier}
	\put(0,23){\begin{rotate}{90}{Accuracy}\end{rotate}}
	\put(40,25){\color{blue} $r$ PCA features}
	\put(40,20){\color{darkgreen} $r$ sensors, no subsampling}
	\put(40,15){\color{red} $r$ sensors from $p=400$ pixels}
	\put(40,10){\color{teal} $r$ random pixels}
	\put(-4,62){(a)}
\end{overpic}  \\
\vspace{25pt}
\begin{overpic}[width=0.43\textwidth]{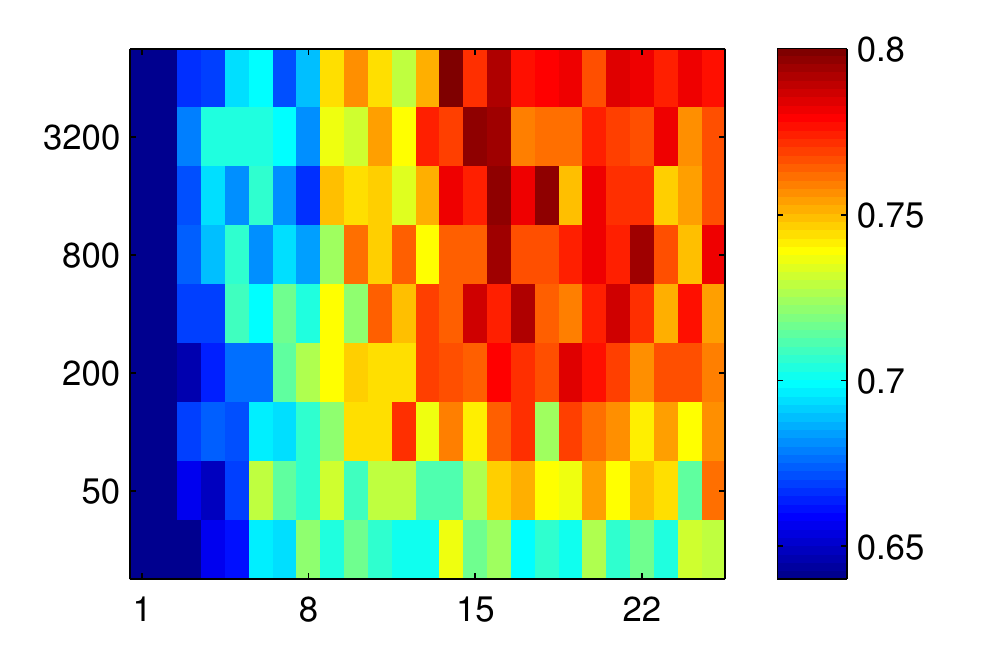}
	\put(40,-3){$r$}
	\put(-2,30){$p$}
	\put(70,2){Mean accuracy}
	\put(-4,62){(b)}
\end{overpic}
\vspace{5pt}
\caption{A cross-validation of classification accuracy between images of cats and dogs.  Panel (a) compares using $r$ learned sensors/features (solid red and green lines) against using $r$ random pixel sensors (dashed blue line) and projections onto the first $r$ principal components of the full image (solid blue line). Each data point summarizes 400 random iterations.  At each iteration, a different 90\% subsample was used to train the classifier, whose accuracy was assessed on the remaining 10\% of images.  Error bars are standard deviations.  (b) A summary of mean cross-validated accuracy varying $p$, the number of pixels used in the random subsample, and $r$, the number of features/sensors used to construct the classifier.
}
\label{fig:errors_per_sensors}
\end{center}
\end{figure}

\begin{figure*}
\begin{center}
\begin{overpic}[width = .9\textwidth]{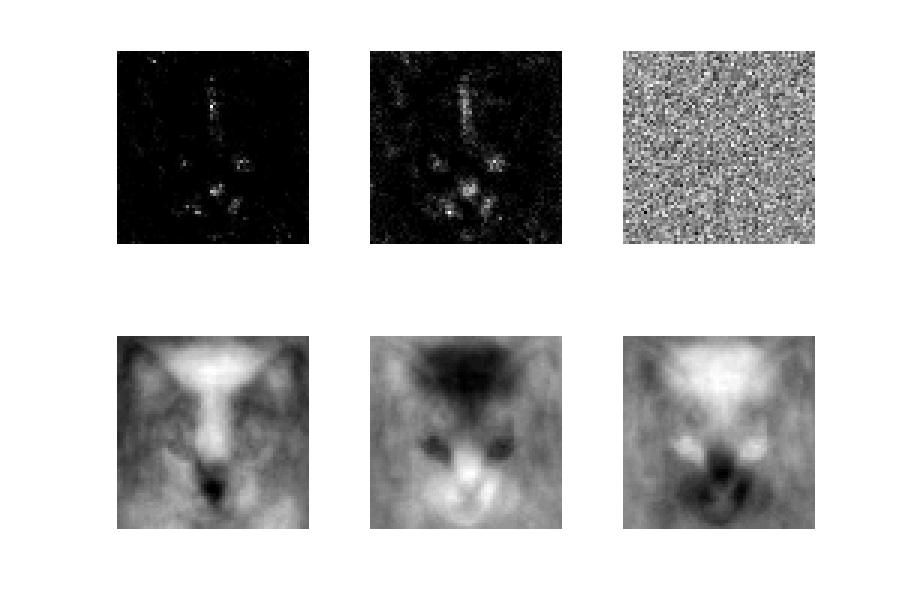}
	\put(11,64){(a)}
	\put(39,64){(b)}
	\put(67,64){(c)}
	\put(11,32){(d)}
	\put(39,32){(e)}
	\put(67,32){(f)}

	\put(16,65){mean sensors}
	\put(15,62){(no subsampling)}

	\put(44,65){mean sensors}
	\put(44,62){($p$ = 400 pixels)}

	\put(72,65){mean sensors}
	\put(72,62){($p$ = 400 r.p.)}

	\put(15,31){mean of raw data}
	\put(46,31){mean cat}
	\put(74,31){mean dog}
\end{overpic}
\vskip -.1in
\caption{The mean sensor locations averaged over 400 random learning iterations (top row), compared to images of the mean cat and the mean dog (bottom row).  For the top row, $r = 15$ sensors locations are learned at each iteration from a random 50\% of images designated as the training set, and the colormap from black to white represents the probability a sensors is at that location.  In other words, we are visualizing the distribution of sensor locations, obtained by summing the rows of either $\bPhihat_1^T$ or $(\bPhihat_2 \bPtilde)^T$.  Panel (a) shows sensors learned with no subsampling; panel (b) shows sensors learned from 400 randomly sampled pixels; panel (c) shows sensors learned from 400 random projections. 
The bottom row shows the centroid of the raw data in panel (c), which was subtracted from each image to obtain $\bX$.  Panels (e) and (f) show the average cat and the average dog after mean subtraction.  Comparing the top row to the difference between (e) and (f), it is apparently that sensors cluster around the forehead, eyes, mouth, and the tops of ears.}
\label{fig:sensor_locations}
\end{center}
\end{figure*}
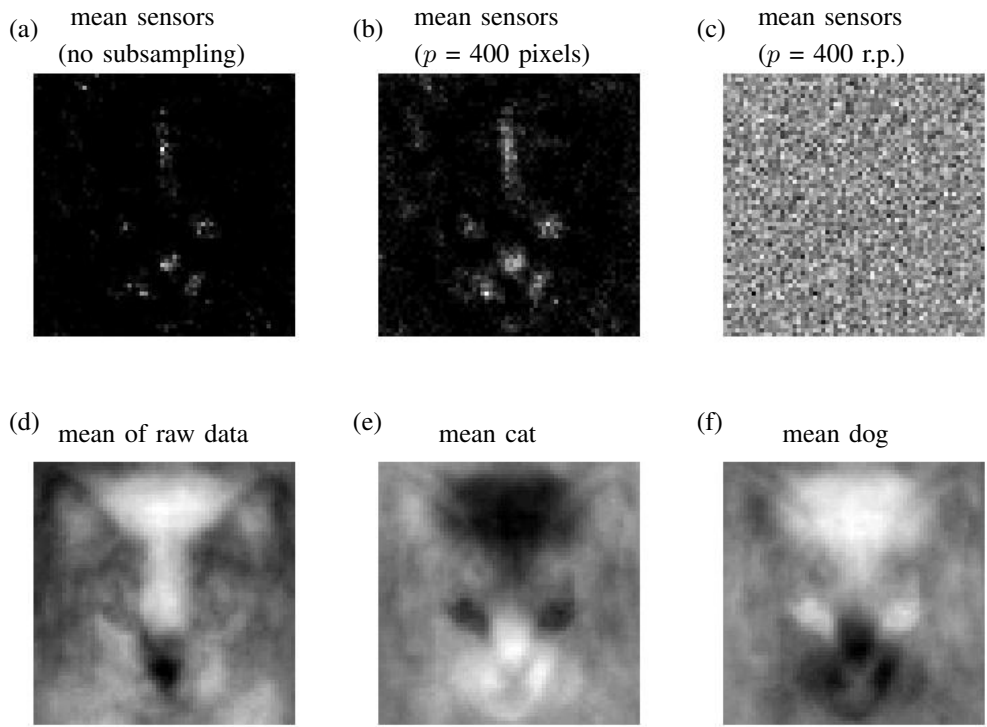

%%% EXAMPLE 1
\subsection{Experiment 1 -- Cats and Dogs}\label{sec:results:catdog}

\begin{figure*}
  \begin{minipage}{0.35\textwidth}
    \centering
    \begin{overpic}[width=1\textwidth]{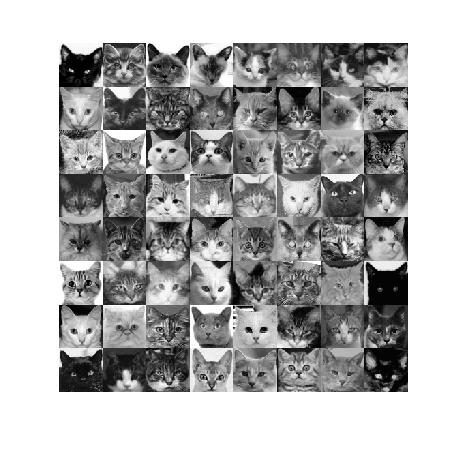} 
    \put(0,95){(a)}
    \end{overpic}
  \end{minipage}\hspace{-20pt}
    \begin{minipage}{0.35\textwidth}
    \centering
    \includegraphics[width=1\textwidth]{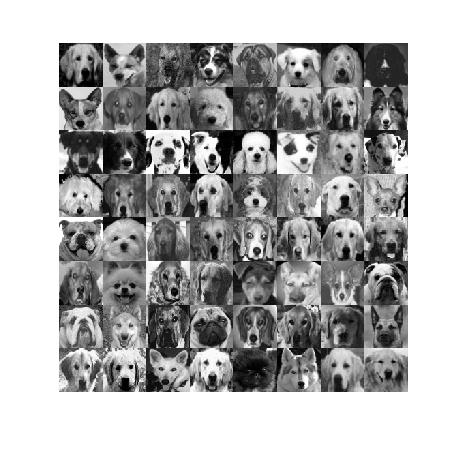} 
  \end{minipage}\hspace{-10pt}
  \begin{minipage}{0.30\textwidth}
    \centering
	\begin{overpic}[width=0.48\textwidth]{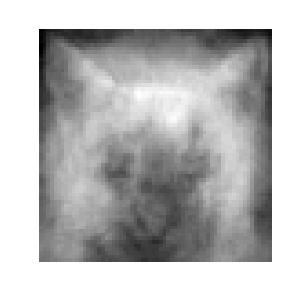}
	\put(0,110){(b)}
	\end{overpic}
	\includegraphics[width=0.48\textwidth]{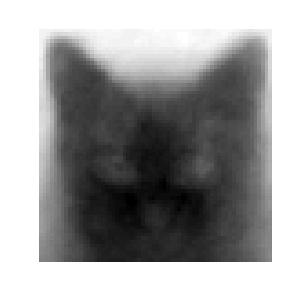}\\
	\includegraphics[width=0.48\textwidth]{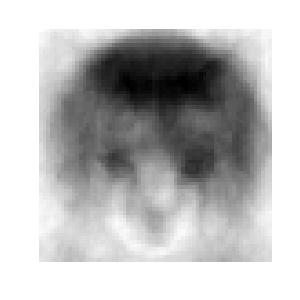} 
	\includegraphics[width=0.48\textwidth]{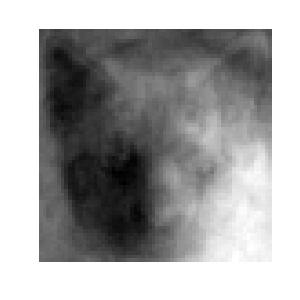}
  \end{minipage}\hfill
  \vskip -.2in
\caption{(a) Examples of cats and dogs images used in the dataset.  Each image has $n = 64 \times 64 = 4096$ pixels. (b) The four (4) PCA modes with the largest singular values of the full image dataset (eigenpets).  Note that the first two modes resemble cats, the third mode resembles a dog, and the next modes are mixtures of cats and dogs.}
\label{fig:all_cats_dogs}
\end{figure*}

Our first example seeks to classify images of cats versus images of dogs.  
Each image belongs to a distinct species, and there is considerable variability between individuals of the same species (Fig.~\ref{fig:all_cats_dogs}).  Notably, members of each species take on a large range of colorations in fur, markings, and ear postures.  We used 121 images each of cats and dogs; images have $n = 64 \times 64 = 4096$ pixels and are stacked into columns of $\bX$.  Figure~\ref{fig:all_cats_dogs}\,(b) shows the first four PCA modes of $\bX$.

We compared classification accuracy using learned measurements as well as using the principal components of the full image and using random pixel measurements.  The accuracy for all of these methods improved with larger numbers of sensors/features; $r$ learned sensors performed almost as well as $r$ principal components and consistently better than $r$ random pixel sensors.   Figure~\ref{fig:errors_per_sensors} shows the results of experiments where classifiers were trained on a random 90\% of the images and assessed on the remaining 10\% of images.  The mean and standard deviation of the cross-validated accuracy over 400 random iterations of training/test images are plotted.  Classifiers built on less than 1\% of the total pixels (green and red lines in Fig.~\ref{fig:errors_per_sensors}\,(a)) performed nearly as well as projections to principal components of the full image.  

Accuracy reached a plateau at around $r=15$ features/sensors (Fig.~\ref{fig:errors_per_sensors}\,(b)) of around 80\%, suggesting the discriminant vector between cats and dogs may be over-fit when more than 15 features were used.  Figure~\ref{fig:errors_per_sensors}\,(b) shows mean classification performance using $r$ sensors learned from $p$ subsampled measurements and $r$ features, where $p$ and $r$ are varied systematically.
Sparse sensors learned from $p=400$ pixels did almost as well as those learned from $p = n = 4096$ pixels.  In other words, we were able to learn a nearly optimally sparse set of sensors starting from an already massively under-sampled set of pixels. 

The performance of all of these approaches is limited by the large variability in appearances within the categories of cats and dogs.  In fact, the images that are most often mis-classified include animals that are predominantly one color (black or white) and dogs with pointy as opposed to droopy ears.  In other words, some dog images may not lie close to the bulk of the other dog images in feature space, so that a linear classifier is not able to capture the complex geometry separating the two categories.

It is possible to build classifiers from random pixels that perform significantly better than chance.  The improvement in accuracy of using computed features or learned sensors over random pixels became smaller as $r$ increases (Fig.~\ref{fig:errors_per_sensors}\,(a)).  The perhaps surprising performance of random pixels is consistent with the notion of enhanced sparsity and Wright \emph{et al.}'s~\cite{Wright:2009} observation that using random features, as long as there is enough of them, serves classification as well as using engineered features.

If the sparse sensors represent pixels that are key to the decision, then they should be clustered at locations that are maximally informative about the difference between cats and dogs.  Indeed, an average of sensor locations learned over 400 random iterations shows clustering around the animals' mouth, forehead, eyes, and tops of ears (Fig.~\ref{fig:sensor_locations}\,(a)).  When sparse sensors were learned from an already subsampled dataset, a qualitatively similar distribution of sensor locations was found, where each cluster of sensors showed a slightly larger spatial variance (Fig.~\ref{fig:sensor_locations}\,(b)).  Such an ensemble of sensor placements can be thought of as a mask for facial features particularly relevant for the classification.

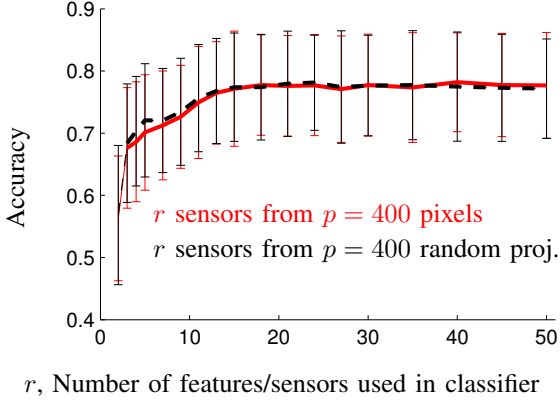
\begin{figure}
\begin{center}
\begin{overpic}[width=0.43\textwidth]{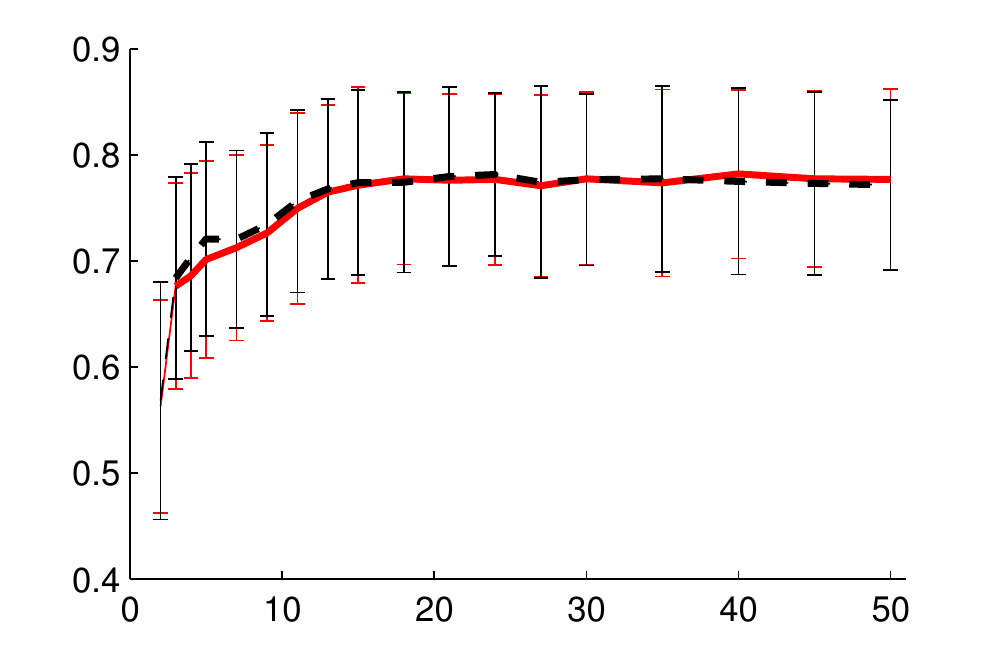} 
	\put(0,-5){$r$, Number of features/sensors used in classifier}
	\put(0,23){\begin{rotate}{90}{Accuracy}\end{rotate}}
	\put(22,24){\color{red} $r$ sensors from $p=400$ pixels}
	\put(22,18){\color{black} $r$ sensors from $p=400$ random proj.}
\end{overpic}
\vspace{25pt}
\caption{Classification accuracy of sensors learned from random projections (black dashed line) is the same as sensors learned from single pixels (red line).  The experiments performed were identical to those show in Fig.~\ref{fig:errors_per_sensors}.  Random projections were ensembles of Bernoulli random variables with mean of 0.5.}
\label{fig:randproj}
\end{center}
\end{figure}

Our algorithm applies equally well when the subsampled dataset (middle row in Fig.~\ref{fig:schematic}) is not a set of random pixels but a set of random projections.  Figure~\ref{fig:randproj} shows that classifiers built on $r$ sensors learned from $p = 400$ pixels or $p = 400$ random projections perform identically.  Random projections used to produce this result comprised of ensembles of Bernoulli random variables with mean of 0.5.  Even so, Fig.~\ref{fig:sensor_locations}\,(c) makes clear that using random projections instead of pixels is a poor engineering choice.  The ensemble of sensors computed from random projections does not illuminate coherent facial features.

%%% EXAMPLE 2
\subsection{Experiment 2 -- Yale B Faces}\label{sec:results:yalefaces}

\begin{figure*}
\hspace{-10pt}
  \begin{minipage}{0.25\textwidth}
    \centering
    \begin{overpic}[width=1\textwidth]{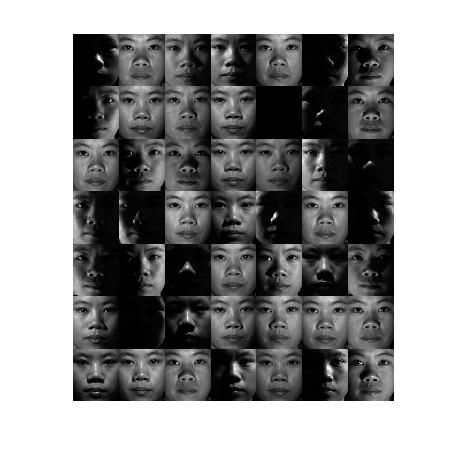} 
    \put(5,95){(a)}
    \end{overpic}
  \end{minipage}\hspace{-15pt}
    \begin{minipage}{0.25\textwidth}
    \centering
    \includegraphics[width=1\textwidth]{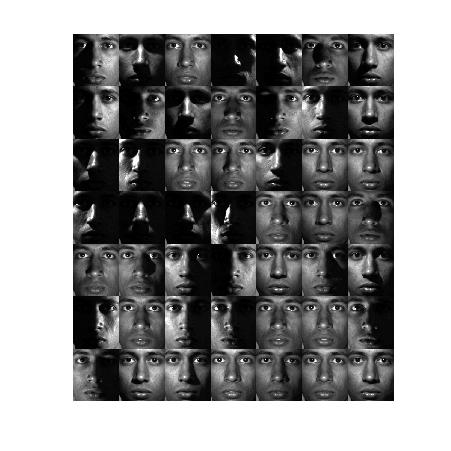} 
  \end{minipage}\hspace{-15pt}
    \begin{minipage}{0.25\textwidth}
    \centering
    \includegraphics[width=1\textwidth]{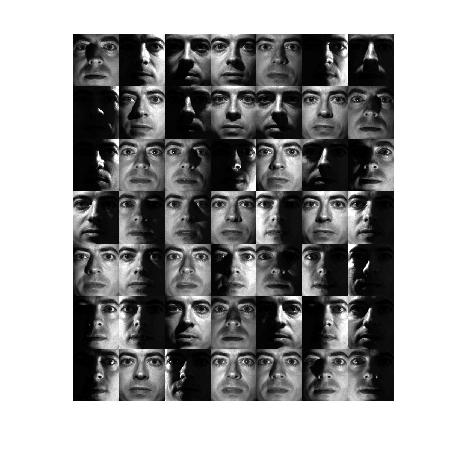} 
  \end{minipage}\hspace{-10pt}
  \begin{minipage}{0.35\textwidth}
    \centering
	\begin{overpic}[width=0.3\textwidth]{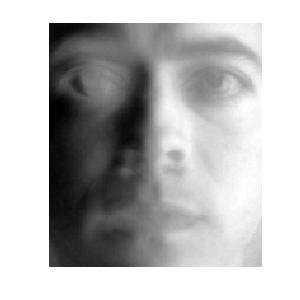}
	\put(0,110){(b)}
	\end{overpic}
	\includegraphics[width=0.3\textwidth]{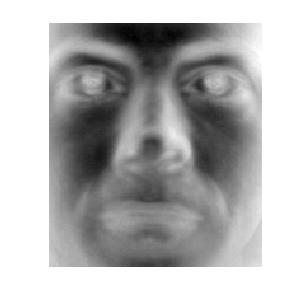}
	\includegraphics[width=0.3\textwidth]{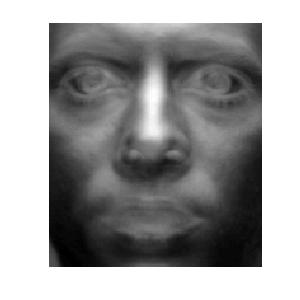} \\
	\includegraphics[width=0.3\textwidth]{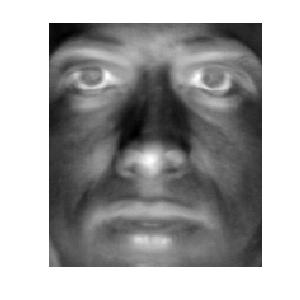}
	\includegraphics[width=0.3\textwidth]{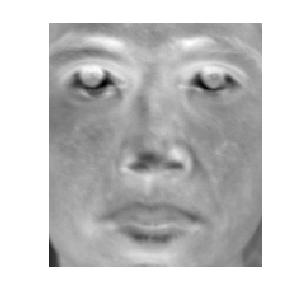}	
	\includegraphics[width=0.3\textwidth]{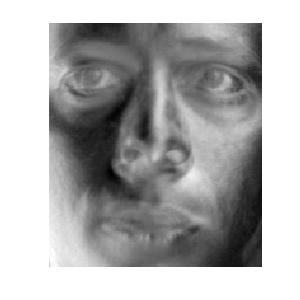}	
  \end{minipage}\hfill
\caption{(a) Examples of images from the Yale Faces Database B.  Different light conditions introduce occlusions to the faces.  There were 64 images per person; each image has $n = 192 \times 168 = 32,256$ pixels. (b) The six (6) PCA modes with the largest singular values of the full image dataset (eigenfaces).}
\label{fig:yaleBfaces}
\end{figure*}

We extended sparse sensor learning to classification between more than two categories, applying our approach to human face recognition.  We used the Yale Faces Database B extended~\cite{georghiades:2001,Lee:2005fk}, which contains images of individual faces captured under various lighting conditions (example images in Fig.~\ref{fig:yaleBfaces}\,(a)).  Images have been perviously aligned and cropped; each has $n = 192 \times 168 = 32,256$ pixels.  Figure~\ref{fig:yaleBfaces}\,(b) shows the first six eigenfaces corresponding to the three example sets of faces.  Comparing the faces dataset with the cat/dog dataset in Sec.~\ref{sec:results:catdog}, there is significantly less variability within each category in facial features, although large portions of the faces were effectively occluded by lack of illumination.  

\begin{figure*}[t]
\begin{center}
\vskip -.1in
\begin{overpic}[width=0.32\textwidth]{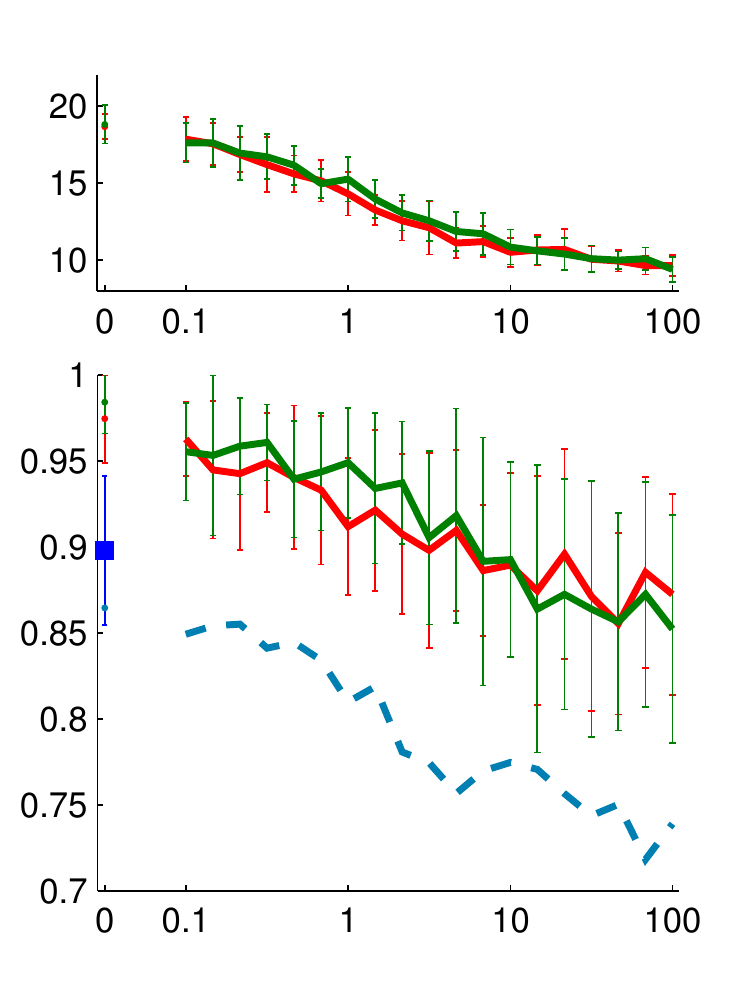}
	\put(0,96){(a)}
	\put(40,0){$\lambda$}
	\put(-1, 70){\begin{rotate}{90}{\# Sensors}\end{rotate}}
	\put(-1, 27){\begin{rotate}{90}{Accuracy}\end{rotate}}
	\put(10,95){$c=3$ Faces, $r=10$ Features}
\end{overpic}
\begin{overpic}[width=0.32\textwidth]{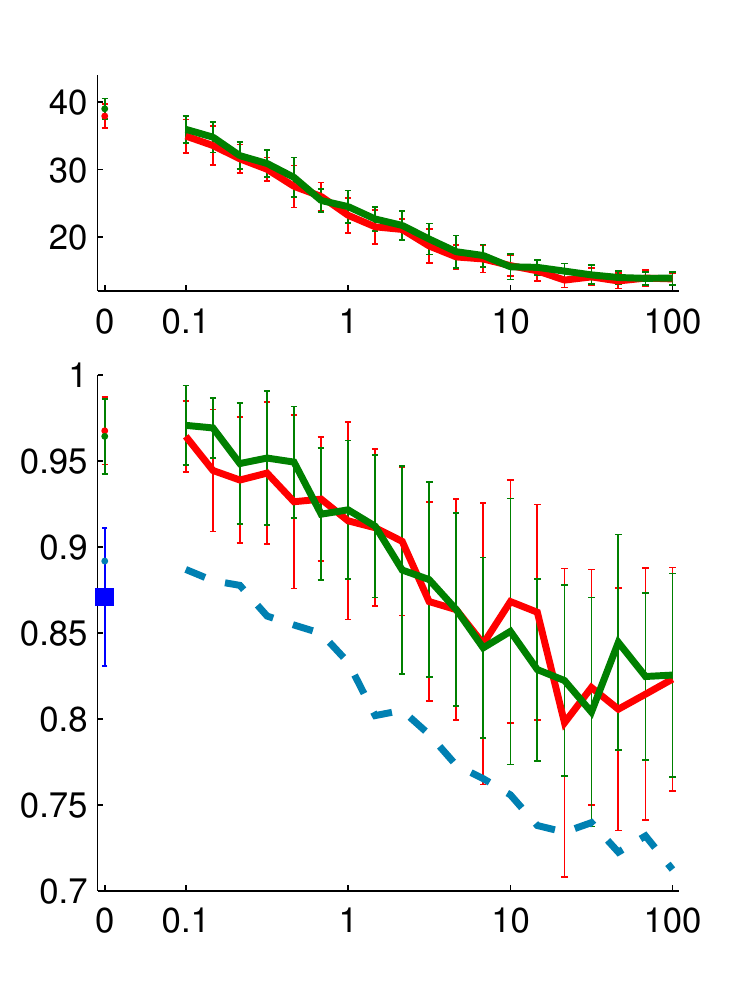}
	\put(0,96){(b)}
	\put(40,0){$\lambda$}
	\put(10,95){$c=4$ Faces, $r=14$ Features}
\end{overpic}
\begin{overpic}[width=0.32\textwidth]{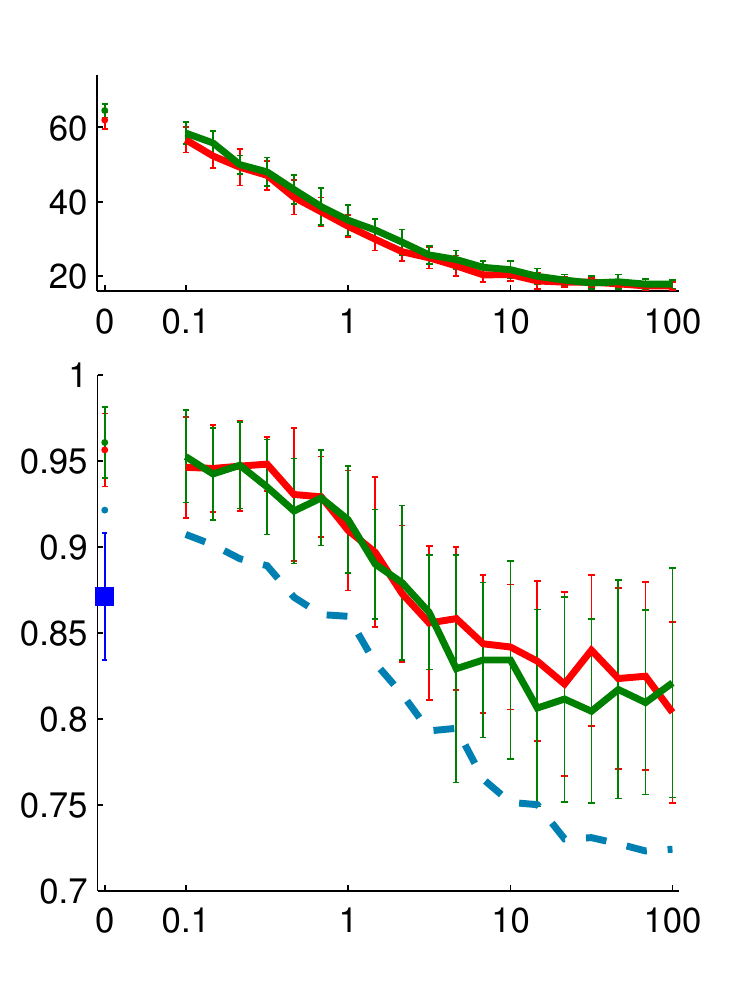}
	\put(0,96){(c)}
	\put(40,0){$\lambda$}
	\put(10,95){$c=5$ Faces, $r=18$ Features}
	\put(42,60){\color{darkgreen} no subsampling}
	\put(42,55){\color{red} $10\times$ subsampled}
	\put(42,50){\color{teal} random pixels}
	\put(13,35){\color{blue}$r$ PCA}
	\put(13,30){\color{blue}features}
\end{overpic}
\vskip .125in
\caption{Classification of 3, 4, and 5 faces as the coupling weight $\lambda$ varies between 0 and 100.  Number of sensors and the cross-validated performance are shown, comparing sensors learned with no subsampling (solid green lines), sensors learned from 1/10 of the pixels (solid red lines), and random sensors of the same number (dashed blue lines) against the accuracy by using $r$ PCA features of the full images (blue square).  Each instance of the classifier was trained on a random 75\% of the images and evaluated on the remaining 25\%; error bars are standard deviations.}
\vskip -.1in
\label{fig:errors_per_lambda}
\end{center}
\end{figure*}
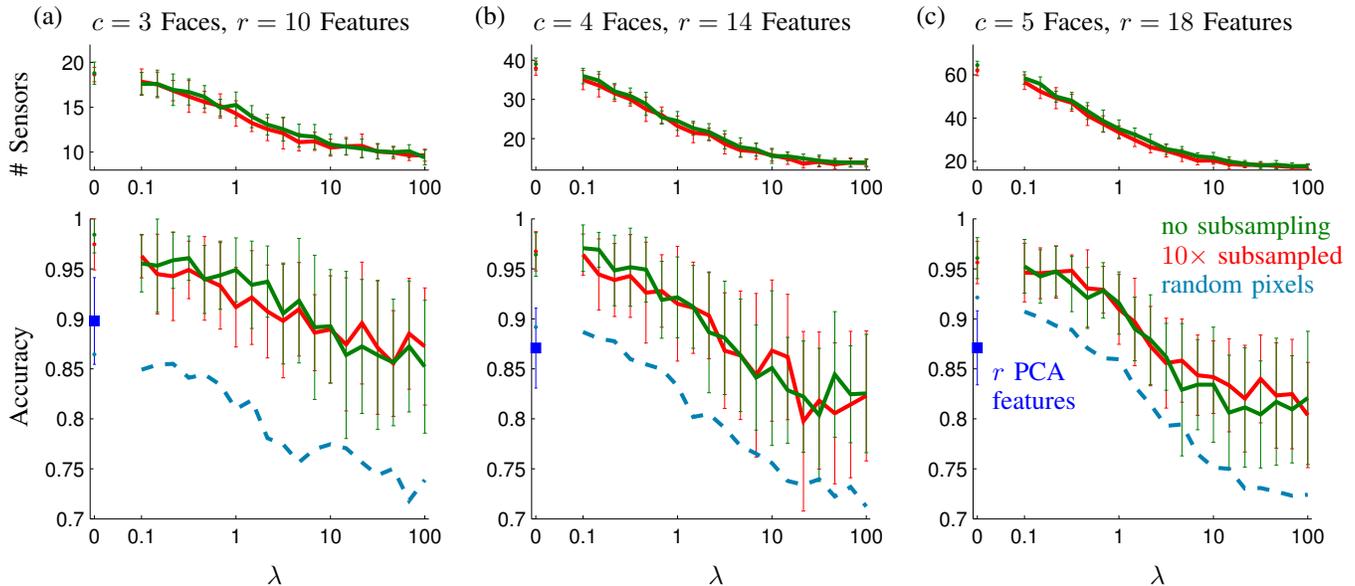

We demonstrated our sensor learning approach to learn sensor locations that categorize $c~=$~3, 4, and 5 faces and compared the classification accuracy of learned sensors to projections to PCA features of the full image and to random sensors of the same number.  Figure~\ref{fig:errors_per_lambda} shows that, as the coupling weight $\lambda$ is increased, the number of learned sensors decreases.  When $\lambda = 0$, an upper bound of $r(c-1)$ pixels locations were found.  In the limit $\lambda \rightarrow \infty$, the number of sensors were bounded by $r$, the number of PCA features included in the discrimination.  By adjusting the magnitude of $\lambda$, we gain control over the number of sensor locations used in the classification.  When individual sensors are expensive, it may be desirable to use fewer sensors at the cost of slightly lower performance.

The cross-validated accuracy shown in the bottom row of Fig.~\ref{fig:errors_per_lambda} are results obtained from re-computing the LDA projection $\hat{\bw}^T$ to decision space after learning the sensor locations.  This strategy is not typically an expensive computation, as the number of rows in the measurement vectors is small.  Further, a LDA classifier tailor-made for the specific learned pixels can out-perform the PCA approach when the number of sensors exceeds the number of PCA features.  In contrast, the induced projection method described in Sec.~\ref{ss:ldaprojection} cannot perform better than the PCA-LDA approach.  For classification between $c > 2$ categories, especially for larger $\lambda$ coupling weights, re-computing the LDA projection usually leads to more accurate results.

Interestingly, sensors learned from randomly subsampled pixels (10\% of the original pixels) performed exactly as well as sensors learned from the full image (red and green lines in Fig.~\ref{fig:errors_per_lambda}).  
Obtaining sparse sensors from already subsampled data presents significant savings in the learning procedure, both in the SVD to extract a feature space and in the convex optimization to solve for a sparse $\bs$.

\begin{figure}
\begin{center}
\vskip -.1in
\begin{overpic}[width=0.24\textwidth]{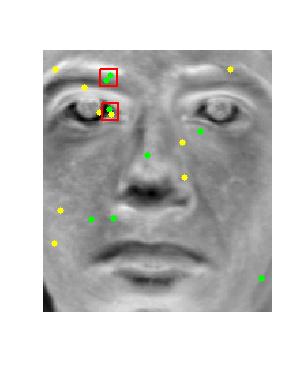}
	\put(0,90){(a)}
	\put(35, 91){$\lambda=0$}
\end{overpic}
\begin{overpic}[width=0.24\textwidth]{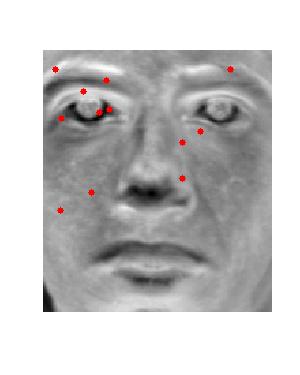}
	\put(0,90){(b)}
	\put(30, 91){$\lambda=0.29$}
\end{overpic}
\vskip -.15in
\caption{Increasing the coupling weight $\lambda$ results in fewer sensor locations that capture approximately the same features.  In both panels, the underlying image is a visualization of the decision vector $\chi = \bPsi_r \bw$ for an example of categorizing $c=3$ faces using $r=10$ features.  Panel (a) shows the locations of sparse sensors that independently reconstruct the $c-1 = 2$ columns of $\bw$ (yellow and green dots, 20 total).  The sensors boxed in red are aggregated in  panel (b), where the coupling weight $\lambda$ brought the total number of sensors down to 15.  }
\vskip -.15in
\label{fig:sensor_locations_lambda}
\end{center}
\end{figure}

Figure~\ref{fig:sensor_locations_lambda} illustrates an example of increasing $\lambda$ on the number and locations of sensors identified by the solution to the optimization in Eq.~\eqref{eq:ell1extended}.  
For this example, $c=3$ faces were in the training set so $\bw$ and $\bs$ each has $c-1 = 2$ columns.
When $\lambda=0$, the columns of $\bw$ (linear discriminant vectors in $r=10$ dimensional PCA feature space) are treated independently, and $r(c-1)=20$ total sensors are found.  
Notice, however, in Fig.~\ref{fig:sensor_locations_lambda}\,(a) that certain pairs of sensor locations are in close proximity (red boxes) and likely carry information about the same non-local facial feature.
As $\lambda$ increases and the total number of sensors is penalized, these sensor pairs appear to collapse onto single sensors in Fig.~\ref{fig:sensor_locations_lambda}\,(b).  
Comparing the two sets of sensor locations, we can see that, in additional to aggregated sensors, some sensors have remained the same (for example, the tops of each eyebrow), some sensors have disappeared (on the cheek, at bottom right), while some entirely new sensors have appeared (lower right corner of eye).

\begin{figure}
\begin{center}
\begin{tabular}{cc}
\begin{overpic}[width=0.24\textwidth]{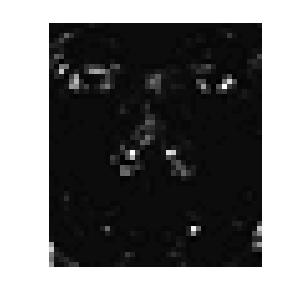}
	\put(0,90){(a)}
	\put(29,96){mean sensors}
\end{overpic} & 
\begin{overpic}[width=0.24\textwidth]{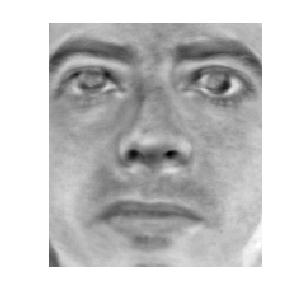}
	\put(0,90){(b)}
	\put(29,96){mean face 1}
\end{overpic} \\
\begin{overpic}[width=0.24\textwidth]{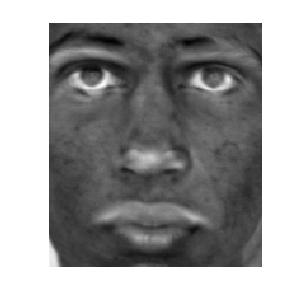}
	\put(0,90){(c)}
	\put(29,96){mean face 2}
\end{overpic}&
\begin{overpic}[width=0.24\textwidth]{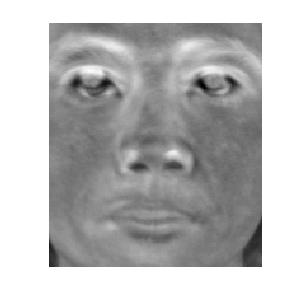}
	\put(0,90){(d)}
	\put(29,96){mean face 3}
\end{overpic}
\end{tabular}
\vskip -.1in
\caption{Mean sensor locations (top left) to discriminate between 3 faces (mean faces b-d), averaged over 400 learning iterations where a random 75\% subset of the images was used as the training set.  The mean sensor locations map was shrunk by a factor of 4 (using a cubic kernel) to emphasize the sensors converging around the eyes, nose, corner of mouth, and arches of eyebrows.}
\label{fig:sensor_locations_3faces}
\end{center}
\end{figure}

Sparse sensors for face recognition cluster at major facial features: the eyes, the nose, and corners of the mouth (Fig.~\ref{fig:sensor_locations_3faces}).  
The nose is particularly prominent in these masks, consistent with the fact that, due to the eccentricity in illumination, the nose is the only facial feature reliably visible in all images.  
Interestingly, this data-driven algorithm identifies the same features that are favored by humans.  
As first noted by Yarbus~\cite{yarbus1967eye}, humans examining an image of a face spend a preponderance of time fixating at the eyes, the nose, and the mouth.

%%%%%%%%%%%%
%%% DISCUSSION
%%%%%%%%%%%%
\section{Discussion}\label{sec:discussion}
\begin{figure*}
\begin{center}
\begin{overpic}[width=.97\textwidth]{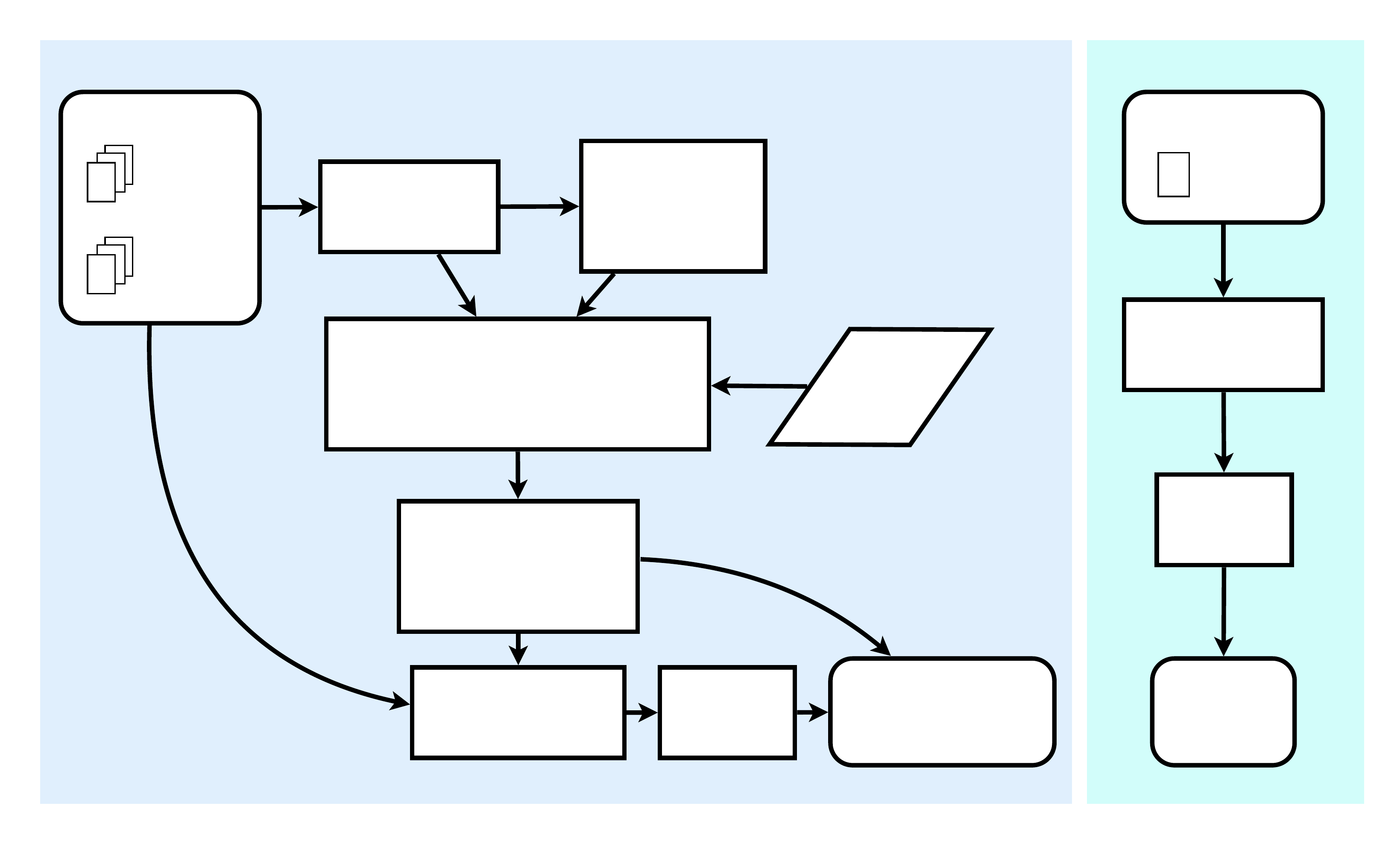}
\put(4,55){(a) Learning Module}
\put(5.5,51){Training Data}
\put(11.5,47){class 1}
\put(14,43){$\vdots$}
\put(11.5,40.5){class $c$}

\put(26,46){Extract}
\put(26,43.5){Features}
\put(37.5,46){$\bPsi$}
\put(30,38.5){$\bPsi$}

\put(43,47){Compute}
\put(43,44.5){Discriminant}
\put(43,42){Directions}
\put(43.5,38.5){$\bw$}

\put(25,34.5){Select Optimally Few Sensors}
\put(28,32){by $\ell_1$-Minimization,}
\put(29,29.5){Eqs.~\eqref{eq:ell1} and \eqref{eq:ell1extended}}
\put(38,26){$\bs$}

\put(61,34.5){Coupling}
\put(59.5,32){Weight}
\put(60,29.5){$\lambda$}

\put(31,21.5){Construct}
\put(31,19){Measurement}
\put(31,16.5){Matrix}
\put(38,13){$\bPhi$}
\put(55,19){$\bPhi$}

\put(31,10){Measure at}
\put(31,7.5){Sparse Sensors}

\put(48,10){Build}
\put(48,7.5){Classifier}

\put(60.5,10){Sensor Locations}
\put(60.5,7.5){ \& Classifier}

\put(78,55){(b) Execution Module}
\put(82,51){Test Data}
\put(87,47){class ?}

\put(81.5,36.5){Measure at}
\put(81.5,34){Sparse Sensors}

\put(84,22.5){Classify}

\put(84,9){Category}
\end{overpic}
\caption{A summary of the two modules that comprise the classification framework.  The image classification procedure may be extended to a broader class of datasets.  Single-pixel measurements become individual sensors in a sensor network.}
\label{fig:extension}
\end{center}
\end{figure*}
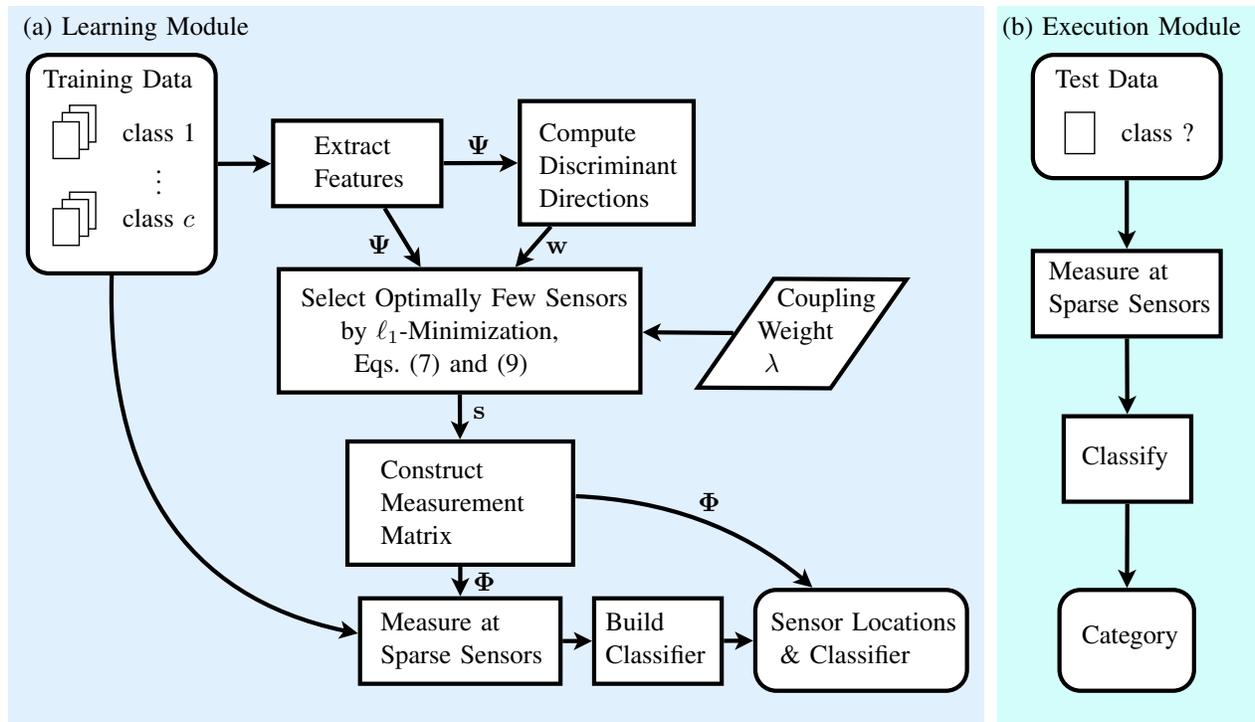

In this paper, we described an algorithm that refines a large set of measurement locations to learn a much smaller subset of key locations to best serve a classification task.  Steps of the algorithm are shown in Fig.~\ref{fig:extension}.
The algorithm exploits \emph{enhanced sparsity} for classification, when reconstruction may be bypassed.  
Enhanced sparsity provides an orders-of-magnitude reduction in number of measurements required when compared with standard compressive sensing strategies.  
Measurements are projected directly into decision space; reconstruction from so few measurements is neither possible or needed.

Our algorithm leverages the sparsity promoting $\ell_1$ minimization as phrased in Eqs.~\eqref{eq:ell1} or \eqref{eq:ell1extended}.  
These convex optimization problems solve for a sparse set of sensor locations to closely approximate the linear discrimination vector in PCA space.
The approach may be generalized to a variety of other dimensionality reduction and discrimination algorithms.  
In addition, once a set of sensors have been obtained, a number of more sophisticated classifier algorithms may be applied to these measurements to improve the classification accuracy.  

We demonstrated our approach on two examples of image recognition.  
Classifiers built on learned sensors approached the performance of classifiers built on projections to principal components of the full images.  
These optimal sensor locations could be learned approximately even from already randomly subsampled images.  
In either case, the learned sensors performed significantly better than the matched number of randomly chosen sensors.  
Further, the ensemble of learned sensor locations clustered around coherent features of the images.  
It is possible to think of this ensemble as a pixel mask for the faces in the training set, which may be of use when applied to engineered or biological systems where the important features may not be salient by inspection.

It is plausible that biological organisms may exploit enhanced sparsity for classification.  
Organisms interact with the external world with motor outputs, which are often discrete trajectories at specific moments in time, so that the transformation from sensory inputs to motor outputs can be thought of as a classification task.  
For example, a fly has no need to reconstruct the full flow velocity field around its body---it has only to decide what to do in response to a gust.  
Sensory organs and data processing by the nervous system can be expensive; therefore, it is advantageous to place a smaller number of sensors at key locations on the body.  

Although introduced for image discrimination, the algorithm may be applied to a variety of non-image data types with more general sensor networks.
%; this is shown schematically in Fig.~\ref{fig:extension}.  
One such application is the detection of disease spread in epidemiological monitoring.  
We also envision these methods applying to mobile sensor networks in oceanographic and atmospheric sampling, as well as to detect and monitor various network behaviors in the electrical grid, internet packet routing, and transportation.

%%%%%%%%%%%%
%%% ACKNOWLEDGEMENTS
%%%%%%%%%%%%
\section*{Acknowledgements}
We especially thank Tom Daniel and Leifeng Bo for helpful discussions related to this work, and Anna Gilbert for insightful comments on the manuscript.
B. W. Brunton and J. N. Kutz  acknowledge support from the National Science Foundation (DMS-1007621); 
S. L. Brunton and J. N. Kutz acknowledge support from the  U.S. Air Force Office of Scientific Research (FA9550-09-0174).  
J. L. Proctor acknowledges support by Intellectual Ventures Laboratory.

%%%%%%%%%%%%%
%%%% APPENDIX
%%%%%%%%%%%%%
%\section*{Appendix}\label{app:datasets}
%In this Appendix, we show examples of the image datasets used in the above manuscript to demonstrate the sparse sensor algorithm described in Sec.~\ref{sec:methods}.
%
%Figure~\ref{fig:all_cats_dogs} shows some of the cat and dog images along with the first four principal components (eigenpets) of the dataset.  
%We used 121 grayscale images each of cats and dogs; each image has $n = 64 \times 64 = 4096$ pixels.  
%Images were chosen to be faces in roughly frontal orientation and were pre-processed by cropping and alignment.
%
%Figure~\ref{fig:yaleBfaces} shows examples of three faces from the Yale Faces Database B~\cite{georghiades:2001,Lee:2005fk}, along with the first six principal components (eigenfaces) of the three individuals in panel (a).  
%The full database contains 64 grayscale images each of 38 individuals, of which we used a random subset as described in the main text.  
%Each image has $n = 192 \times 168 = 32,256$ pixels.  
%Images had been acquired in the frontal orientation under a variety of illumination conditions.  
%When the illumination was particularly eccentric, large portions of the faces were effectively occluded.

%%%%%%%%%%%%
%%% BIBLIOGRAPHY
%%%%%%%%%%%%
\bibliographystyle{plain}
\bibliography{sparse_sensors}

\end{document}